\documentclass[11pt]{article}
\usepackage{custom}

\title{Uncertainty Quantification for Large-Scale Deep Networks via Post-StoNet Modeling}
\author{Yan Sun and Faming Liang}
\date{University of Pennsylvania and Purdue University}

\begin{document}

\maketitle

\begin{abstract}
     Deep learning has revolutionized modern data science. However, how to accurately quantify the uncertainty of predictions from large-scale deep neural networks (DNNs) remains an unresolved issue. To address this issue,
 we introduce a novel post-processing approach. This approach feeds the output from the last hidden layer of a pre-trained large-scale DNN  model into a stochastic neural network (StoNet), then trains 
 the StoNet with a sparse penalty on a validation dataset and constructs prediction intervals 
 for future observations. We establish a theoretical guarantee 
 for the validity of this approach; in particular, the parameter estimation consistency for the sparse StoNet is essential for the success of this approach. 
 Comprehensive experiments demonstrate that the proposed approach can construct honest confidence intervals with shorter interval lengths 
 compared to conformal methods and achieves better calibration compared to other post-hoc calibration techniques. Additionally, we show 
 that the StoNet formulation provides us with a platform to adapt sparse learning theory and methods from linear models to DNNs.  
\end{abstract}

\section{Introduction}

During the past two decades, deep learning has revolutionized modern data science. It has achieved remarkable success in many scientific fields such as computer vision,
protein structure prediction, 
and natural language processing. 
Despite these successes, how to accurately quantify the prediction uncertainty of DNN models remains an unresolved issue. In practice, it has been widely observed that many large-scale DNN models are miscalibrated, resulting in overconfident predictions \citep[see, e.g.,][]{guo2017calibration, minderer2021revisiting}. As large-scale DNN models are increasingly used in safety-critical applications such as self-driving cars \citep{bojarski2016end} and medical diagnosis \citep{esteva2017dermatologist}, it is crucial that these models not only predict accurately but also are equipped with rigorous uncertainty quantification for their predictions.

To address this issue, efforts have been made in multiple directions. One approach is to use the conformal method \citep{Vovk2005AlgorithmicLI}, a general technique for constructing valid prediction sets for black-box machine learning models. However, if the model is not properly trained, such as in cases of overfitting (which often occurs  for large-scale DNNs), the resulting prediction interval can be overly wide. Another line of research focuses on model calibration, aiming to improve calibration either by modifying the training procedure \citep{kumar2018trainable, mukhoti2020calibrating} or by learning 
a re-calibration map on a separate validation dataset to transform the predictions into better-calibrated ones \citep{zadrozny2002transforming, 
guo2017calibration, kumar2019verified}. However, these methods do not provide a theoretical guarantee that the learned model provides calibrated outputs reliable for decision-making.
In principle, Bayesian methods can also be used to address this issue. However, efficiently simulating from the posterior distribution of a large-scale DNN model remains a challenging problem.

\textcolor{black}{
This paper introduces a post-StoNet modeling approach,
a novel post-processing technique for uncertainty quantification of pre-trained large-scale DNN models. 
The new approach models the relationship between the response variables and the features learned by the pre-trained DNN (specifically, the output from its last hidden layer) using a stochastic neural network (StoNet) \citep{LiangSLiang2022}. 
It trains a sparse StoNet on the validation dataset, and constructs prediction intervals for future observations according to Eve's law. A theoretical guarantee for the 
validity of the resulting prediction intervals is provided. Extensive experiments show that the post-StoNet approach yields honest confidence intervals with shorter lengths compared to conformal methods and achieves better calibration than other post-hoc calibration techniques. Moreover, this paper shows that the StoNet formulation bridges sparse learning theory from linear models to deep neural networks. The consistency of parameter estimation for the sparse StoNet is essential for the success of the proposed approach.}

\paragraph{Related Works}
 
The present work is connected to the literature of statistics and machine learning across a few topics:  

\noindent
{\it $\bullet$ Stochastic Neural Networks.} Stochastic neural networks have been explored in the literature from various perspectives. Notable examples include restricted Boltzmann machines \citep{salakhutdinov2009deep} and deep belief networks \citep{hinton2006fast}, which develop probabilistic generative models  composed of multiple layers of latent variables. Another strand of research focuses on techniques that introduce noise during DNN training to enhance model performance, see e.g., dropout methods \citep{srivastava2014dropout,kingma2015variational}. 
This paper utilizes the stochastic neural network model developed in  \cite{LiangSLiang2022}, reformulating the neural network as a latent variable model. As discussed in Section \ref{sect:post-stonet}, this latent variable framework bridges statistical theories from linear models to deep learning, facilitating uncertainty quantification of 
DNN predictions.

\noindent {\it $\bullet$ Sparse Deep Learning.} Sparse deep learning typically aims to reduce the number of parameters of a DNN while preserving its prediction accuracy. Extensive research has focused on parameter pruning \cite[see e.g.,][]{han2015learning, louizos2017bayesian,
Ghosh2018Structured}. Theoretical properties of sparse DNN models have also been studied in the literature. For instance, \cite{bolcskei2019optimal} quantified the minimum network connectivity required for uniform approximation rates across a class of affine functions, 
and \cite{SunSLiang2021, sun2022learning} established posterior consistency for sparse Bayesian neural networks employing a mixture-Gaussian or spike-and-slab prior. Additionally, \cite{sun2021sparse} and \cite{wang2020uncertainty} explored posterior normality (Bernstein–von Mises theorem) for functionals of sparse neural networks. We show that the StoNet provides a general framework for studying sparse DNN models with various penalty functions.  

\noindent {\it $\bullet$ Calibration.} Many popular deep learning models are reported to be poorly calibrated \citep{guo2017calibration}, prompting a significant body of research focused on enhancing the calibration of machine learning methods \cite[see e.g.,][]{
kumar2018trainable, 
mukhoti2020calibrating}. We address this challenge by offering a post-hoc calibration technique with theoretical guarantees. 

\noindent {\it $\bullet$ Prediction Set.} Conformal prediction methods have gained popularity in recent years as general approaches for constructing valid prediction sets for black-box machine learning models. Full conformal \citep{Vovk2005AlgorithmicLI} or jackknife+ \citep{barber2021predictive} can be computationally expensive. Split conformal, on the other hand, relies on a separate validation set to compute non-conformity scores. 
This study demonstrates that by utilizing a simple sparse StoNet  trained on the validation set, we can effectively construct honest  and more compact prediction sets.

The remaining part of the paper is organized as follows.
Section \ref{stonetsection} provides a brief review for the theory
of StoNets. 
Section \ref{sect:post-stonet} presents the proposed approach and the 
theoretical guarantee for its validity.
Section \ref{simulationsection} illustrates the proposed approach using a simulated example.
Section \ref{sect:examples} presents numerical results on some benchmark datasets. Section \ref{sect:discussion} concludes the paper with a brief discussion.

\section{Asymptotic Equivalence between StoNet and DNN} \label{stonetsection}

This section provides a brief review of the StoNet model and theory regarding asymptotic equivalence between the StoNet and DNN models, which was originally  
established in \cite{LiangSLiang2022}.

Consider a DNN model with $h$ hidden layers, where, for simplicity, assume each hidden neuron has the same activation function. For a regression problem, by separating the feeding and activation operations of each hidden neuron, the DNN model can be written as:
\begin{equation}\label{eq:DNN}
\small
    \begin{split}
    \tilde{\bY}_1 &= \bb_1 + \bw_1\bX, \\ \tilde{\bY}_i & = \bb_i + \bw_i\Psi(\tilde{\bY}_{i-1}), \quad i=2,3,\dots,h,\\
    \bY &= \bb_{h+1} + \bw_{h+1}\Psi(\tilde{\bY}_{h}) + \be_{h+1}, 
    \end{split}
\end{equation}
where $\be_{h+1}\sim N(0, \sigma_{h+1}^2 I_{d_{h+1}})$ is the Gaussian random error; 
$d_i$ denotes the width of layer $i$ for $i=1,2,\ldots,h$ and $d_0 = p$ denotes the dimension of $\bX$;
$\bw_i \in \mathbb{R}^{d_i \times d_{i-1}}, \bb_i \in \mathbb{R}^{d_i}$ denote the weights and bias of $i$-th layer and $\tilde{\bY}_i \in \mathbb{R}^{d_i}$ denotes the pre-activation;
$\Psi(\cdot)$ denote the element-wise activation function s.t. $\Psi(\tilde{\bY}_{i-1})=(\psi(\tilde{Y}_{i-1,1}), \psi(\tilde{Y}_{i-1,2}), \ldots \psi(\tilde{Y}_{i-1,d_{i-1}}))^T$. 
For classification problems, the third equation of (\ref{eq:DNN}) can be replaced with a logit model. 

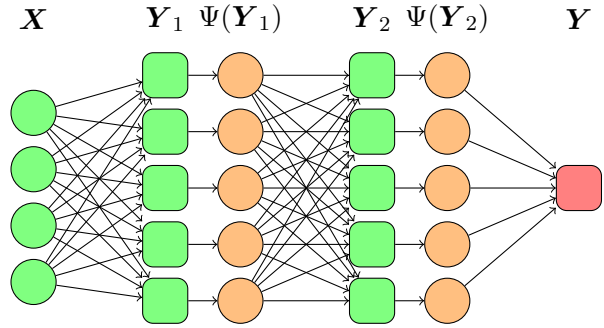
\begin{figure}[htbp]
\begin{center}
\begin{tikzpicture}[node distance=0.75cm]
\tikzstyle{randomnode} = [rectangle, rounded corners, minimum width=0.6cm, minimum height=0.6cm,text centered, draw=black, fill=blue!50]
\tikzstyle{inputnode} = [circle, minimum width=0.6cm, minimum height=0.6cm, text centered, draw=black, fill=green!50]
\tikzstyle{hiddennode} = [circle, minimum width=0.6cm, minimum height=0.6cm, text centered, draw=black, fill=orange!50]
\tikzstyle{outputnode} = [rectangle, rounded corners, minimum width=0.6cm, minimum height=0.6cm,text centered, draw=black, fill=red!50]
\tikzstyle{arrow} = [thick,->,>=stealth]

\node(X1)[inputnode]{};
\node(X2)[inputnode,below of=X1]{};
\node(X3)[inputnode,below of=X2]{};
\node(X4)[inputnode,below of=X3]{};
\node[above of=X1,yshift=0.5cm]{$\bX$};

\node(Y1)[randomnode,right of=X1,xshift=1.0cm,yshift=0.5cm]{};
\node(Y2)[randomnode,right of=X2,xshift=1.0cm,yshift=0.5cm]{};
\node(Y3)[randomnode,right of=X3,xshift=1.0cm,yshift=0.5cm]{};
\node(Y4)[randomnode,right of=X4,xshift=1.0cm,yshift=0.5cm]{};
\node(Y5)[randomnode,right of=X4,xshift=1.0cm,yshift=-0.25cm]{};
\node[above of=Y1,yshift=0.0cm]{$\bY_1$};

\node(Z1)[hiddennode,right of=Y1,xshift=0.25cm]{};
\node(Z2)[hiddennode,right of=Y2,xshift=0.25cm]{};
\node(Z3)[hiddennode,right of=Y3,xshift=0.25cm]{};
\node(Z4)[hiddennode,right of=Y4,xshift=0.25cm]{};
\node(Z5)[hiddennode,right of=Y5,xshift=0.25cm]{};
\node[above of=Z1,yshift=0.0cm]{$\Psi(\bY_1)$};

\node(U1)[randomnode,right of=Z1,xshift=1.0cm]{};
\node(U2)[randomnode,right of=Z2,xshift=1.0cm]{};
\node(U3)[randomnode,right of=Z3,xshift=1.0cm]{};
\node(U4)[randomnode,right of=Z4,xshift=1.0cm]{};
\node(U5)[randomnode,right of=Z5,xshift=1.0cm]{};
\node[above of=U1,yshift=0.0cm]{$\bY_2$};

\node(V1)[hiddennode,right of=U1,xshift=0.25cm]{};
\node(V2)[hiddennode,right of=U2,xshift=0.25cm]{};
\node(V3)[hiddennode,right of=U3,xshift=0.25cm]{};
\node(V4)[hiddennode,right of=U4,xshift=0.25cm]{};
\node(V5)[hiddennode,right of=U5,xshift=0.25cm]{};
\node[above of=V1,yshift=0.0cm]{$\Psi(\bY_2)$};

\node(W)[outputnode,right of=V3,xshift=1.0cm]{};
\node[above of=W,yshift=1.5cm]{$\bY$};
 \draw[->](X1)--(Y1);
 \draw[->](X1)--(Y2);
 \draw[->](X1)--(Y3);
 \draw[->](X1)--(Y4);
 \draw[->](X1)--(Y5);
 \draw[->](X2)--(Y1);
 \draw[->](X2)--(Y2);
 \draw[->](X2)--(Y3);
 \draw[->](X2)--(Y4);
 \draw[->](X2)--(Y5);
 \draw[->](X3)--(Y1);
 \draw[->](X3)--(Y2);
 \draw[->](X3)--(Y3);
 \draw[->](X3)--(Y4);
 \draw[->](X3)--(Y5);
 \draw[->](X4)--(Y1);
 \draw[->](X4)--(Y2);
 \draw[->](X4)--(Y3);
 \draw[->](X4)--(Y4);
 \draw[->](X4)--(Y5);
 
 \draw[->](Y1)--(Z1);
 \draw[->](Y2)--(Z2);
 \draw[->](Y3)--(Z3);
 \draw[->](Y4)--(Z4);
 \draw[->](Y5)--(Z5);
 
 \draw[->](Z1)--(U1);
 \draw[->](Z1)--(U2);
 \draw[->](Z1)--(U3);
 \draw[->](Z1)--(U4);
 \draw[->](Z1)--(U5);
 \draw[->](Z2)--(U1);
 \draw[->](Z2)--(U2);
 \draw[->](Z2)--(U3);
 \draw[->](Z2)--(U4);
 \draw[->](Z2)--(U5);
 \draw[->](Z3)--(U1);
 \draw[->](Z3)--(U2);
 \draw[->](Z3)--(U3);
 \draw[->](Z3)--(U4);
 \draw[->](Z3)--(U5);
 \draw[->](Z4)--(U1);
 \draw[->](Z4)--(U2);
 \draw[->](Z4)--(U3);
 \draw[->](Z4)--(U4);
 \draw[->](Z4)--(U5);
 \draw[->](Z5)--(U1);
 \draw[->](Z5)--(U2);
 \draw[->](Z5)--(U3);
 \draw[->](Z5)--(U4);
 \draw[->](Z5)--(U5);
  
 \draw[->](U1)--(V1);
 \draw[->](U2)--(V2);
 \draw[->](U3)--(V3);
 \draw[->](U4)--(V4);
 \draw[->](U5)--(V5);
 
 \draw[->](V1)--(W);
 \draw[->](V2)--(W);
 \draw[->](V3)--(W);
 \draw[->](V4)--(W);
 \draw[->](V5)--(W);
\end{tikzpicture}
\end{center}
\caption{Illustration of the structure of the StoNet, where each square neuron represents a linear/logistic regression: 
 $\bX$ represents input variable, $\bY_1=\bb_1+\bw_1 X+\be_1$ and $\bY_2=\bb_2+\bw_2 \Psi(\bY_1)+\be_2$ represent latent variables, $\bY=\bb_3+\bw_3 \Psi(\bY_2)+\be_3$ represent output variables, and $\Psi(\cdot)$ represents the activation function.} 
\label{fig:stonet}
\end{figure}

The StoNet, depicted by Figure \ref{fig:stonet}, 
is a {\it probabilistic deep learning model} constructed by introducing auxiliary noise to the $\tilde{\bY}_i$'s in (\ref{eq:DNN}).
 Mathematically, the StoNet is defined as:
\begin{equation}\label{eq:stonet}
\small
 \begin{split} \bY_1 &= \bb_1 + \bw_1\bX + \be_1, \\
 \bY_i &= \bb_i + \bw_i\Psi(\bY_{i-1})+ \be_i, \quad i=2,3,\dots,h, \\ \bY &= \bb_{h+1} + \bw_{h+1}\Psi(\bY_{h}) + \be_{h+1}, 
 \end{split} 
\end{equation}
where the model is composed of a series of simple regressions, 
with $\bY_1$, $\bY_2$, $\ldots$, $\bY_h$ being latent variables. For simplicity, we assume that $\be_i \sim N(0, \sigma_i^2 I_{d_i})$ for $i=1,2,\dots,h,h+1$, although other distributions can also be used. For example, \cite{sunLiang2022kernel} used a modified double exponential distribution for $\be_1$ such that support vector regression (SVR) applies for the first hidden layer. 
In classification tasks, $\sigma_{h+1}^2$ serves as the temperature parameter for the binomial or multinomial distribution at the output layer.
Together with the parameters 
${\sigma_1^2,\ldots,\sigma_h^2}$, 
it regulates the variability of the latent variables ${\bY_1,\ldots,\bY_h}$. 

Let $\btheta = (\bw_1, \bb_1, \dots, \bw_{h+1}, \bb_{h+1}) \in \bTheta$ denote the collection of all parameters of the StoNet, and let 
$\bYmis: = (\bY_1, \dots, \bY_h)$ 
denote the collection of latent variables in  the StoNet.
Let $\pi_{\rm DNN}(\bY|\bX, \btheta)$ denote the likelihood function of the DNN model (\ref{eq:DNN}), and let $\pi(\bY, \bYmis | \bX, \btheta)$ denote the complete data likelihood function of the StoNet (\ref{eq:stonet}).  
To establish the asymptotic equivalence between the StoNet and DNN models,
\cite{LiangSLiang2022} made the following assumptions regarding the network structure, the activation function, and the variances of latent variables.

\begin{assumption} \label{ass:1}
(i) Parameter space $\bTheta$ is compact;
(ii) for any $\btheta \in \bTheta$, $\mathbb{E}(\log\pi(\bY, \bYmis|\bX, \btheta))^2<\infty$ 
; 
(iii) the activation function $\psi(\cdot)$ is Lipschitz continuous with Lipschitz constant $c$; 
(iv) the network's widths $d_l$'s and depth $h$ are allowed to increase with $n$; 
(v) the noise introduced in StoNet satisfies the following condition: $\sigma_{1} \leq \sigma_{2} \leq \cdots \leq \sigma_{h+1}$, $\sigma_{h+1}=O(1)$, and $d_{h+1} (\prod_{i=k+1}^{h} d_i^2) d_k \sigma^2_{k} \prec \frac{1}{h}$ for any $k\in\{1,2,\dots,h\}$.
\end{assumption}

Assumption \ref{ass:1}-(iii) enables the StoNet to accommodate a broad spectrum of Lipschitz continuous activation functions, such as {\it tanh}, {\it sigmoid}, and {\it ReLU}. 
Assumption \ref{ass:1}-(v) limits the variance of noise added to each hidden neuron, where the term $d_{h+1} (\prod_{i=k+1}^{h} d_i^2) d_k$ represents the noise amplification factor of $\be_k$ at the output layer. Generally, the noise introduced in the first few hidden layers should be small to prevent large random errors propagated to the output layer. Under Assumption \ref{ass:1}, part (i) of Lemma \ref{lemma:2} was proven in \cite{LiangSLiang2022}.

Let $Q^*(\btheta)=\mathbb{E}(\log\pi_{\rm DNN}(\bY|\bX,\btheta))$, where the expectation is taken with respect to the joint distribution $\pi(\bX,\bY)$. \textcolor{black}{
Under Assumption \ref{ass:1}-(i)\&(ii) and certain regularity conditions,  Theorem 1 in \citet{Liang2018missing} implies  that 
\begin{equation}\label{eq:sameloss2}
\small
    \frac{1}{n}\sum_{i=1}^n\log\pi_{\rm DNN}(\bY^{(i)}|\bX^{(i)},\btheta)-Q^*(\btheta)\overset{p}{\rightarrow} 0,
\end{equation}
holds uniformly over $\Theta$, where the dimension of $\btheta$ is allowed 
to grow with $n$ at a polynomial rate 
of $O(n^\gamma)$ for some constant 
$0< \gamma < \infty$. This growth rate is typically satisfied by DNNs.} 
Regarding the energy surface of the DNN, they  
assumed $Q^*(\btheta)$ satisfies the following regularity conditions:

\begin{assumption}\label{ass:2}
(i) $Q^*(\btheta)$ is continuous in $\btheta$ and uniquely maximized at $\btheta^*$;
(ii) for any $\epsilon>0$, $sup_{\btheta\in\Theta\backslash B(\epsilon)}Q^*(\btheta)$ exists, where $B(\epsilon)=\{\btheta:\|\btheta-\btheta^*\|<\epsilon\}$, and $\delta=Q^*(\btheta^*)-sup_{\btheta\in\Theta\backslash B(\epsilon)}Q^*(\btheta)>0$.
\end{assumption}

Assumption \ref{ass:2} imposes restrictions on the shape of $Q^*(\btheta)$ near the global maximizer, ensuring it is neither discontinuous nor excessively flat. Given the inherent nonidentifiability of neural network models, Assumption \ref{ass:2} implicitly assumes that each $\btheta$ is unique, subject to loss-invariant transformations such as reordering hidden neurons within the same layer or simultaneously altering the signs or scales of certain connection weights and biases. Under Assumptions \ref{ass:1} and \ref{ass:2}, \cite{LiangSLiang2022} established 
the lemma:

\begin{lemma} \label{lemma:2} \citep{LiangSLiang2022} Suppose that Assumptions \ref{ass:1}-\ref{ass:2}  hold, and $\pi(\bY,\bYmis|\bX,\btheta)$ is continuous in $\btheta$. Then
\begin{equation*} 
\small
 \begin{split}
 (i)  &  \quad  \sup_{\btheta\in \Theta}\Big|\frac{1}{n}\sum_{i=1}^n\log\pi(\bY^{(i)}, \bY^{(i)}_{mis}|\bX^{(i)},\btheta)
 -\frac{1}{n}\sum_{i=1}^n\log\pi_{\rm DNN}(\bY^{(i)}|\bX^{(i)},\btheta)\Big|\overset{p}{\rightarrow} 0, \\
 (ii)  & \quad \|\hat{\btheta}_n-\btheta^*\|\overset{p}{\rightarrow}0, \quad \mbox{as $n\to \infty$},
\end{split}
\end{equation*}
where  $\btheta^*=\arg\max_{\btheta\in \Theta}\mathbb{E}(\log\pi_{\rm DNN}(\bY|\bX,\btheta))$ denotes the true parameters of the DNN model as specified in (\ref{eq:DNN}), and 
\[
\small
\hat{\btheta}_n = \arg\max_{\btheta\in\Theta}\{\frac{1}{n}\sum_{i=1}^n\log\pi(\bY^{(i)},\bY_{mis}^{(i)}|\bX^{(i)},\btheta)\}
\]
denotes the maximum likelihood estimator of the StoNet model (\ref{eq:stonet}) with the pseudo-complete data. 
\end{lemma}

Lemma \ref{lemma:2} suggests that the StoNet and DNN are asymptotically equivalent as the training sample size $n$ grows, establishing the foundation for the bridging property of the StoNet. 
This asymptotic equivalence can be elaborated from two perspectives. First, if the DNN model (\ref{eq:DNN}) is true, Lemma \ref{lemma:2} implies that as $n$ becomes large, the parameters of the DNN can be 
effectively learned by training a StoNet of the same structure,
 provided the $\sigma_i^2$'s satisfy Assumption \ref{ass:1}-(v). 
Conversely,  if the  StoNet (\ref{eq:stonet}) is true,  
Lemma \ref{lemma:2} implies that for any StoNet satisfying Assumptions \ref{ass:1} and \ref{ass:2},  the parameters of the StoNet can be effectively learned by training a DNN of the same structure as $n$ becomes large.

Lemma \ref{lemma:2} further suggests that, similar to the DNN, the StoNet has the universal approximation property for representing probability distributions. Refer to \cite{Lu2020AUA} for the establishment of this property for the DNN. Theoretically, all the DNN approximation properties can be carried over to the StoNet.

\section{Post-StoNet Modeling} \label{sect:post-stonet}

This section presents the theoretical guarantees for sparse StoNets and introduces the post-StoNet approach. We begin by outlining the training algorithm for StoNet in Section \ref{IRO_section}, establish the  consistency for the sparse StoNet in Section \ref{consistency_section}, and describe the method for constructing prediction intervals in Section \ref{sect:CI}. Finally, we formally introduce the Post-StoNet approach in Section \ref{post_stonet_section}.

\subsection{The Imputation Regularized-Optimization Algorithm} 
\label{IRO_section}
By treating the latent variables $\bYmis$ in (\ref{eq:stonet}) as missing data, the StoNet can be trained using the imputation regularized-optimization (IRO) algorithm \citep{Liang2018missing}, a stochastic expectation maximum (EM)-type algorithm specifically tailored for high-dimensional problems. With the IRO algorithm, we 
show that the StoNet provides us with a platform to transfer the theory and methods from linear models to DNNs.

In this subsection, we rewrite the network depth $h$ as $h_n$, rewrite the network widths $(p,d_1,\ldots,d_{h+1})$ as $(p_n,d_{1,n},\ldots, d_{h+1,n})$, rewrite the layer-wise variance $\bsigma^2=(\sigma_1^2,\ldots,\sigma_{h+1}^2)$ as 
$\bsigma_n^2=(\sigma_{1,n}^2,\ldots,\sigma_{h+1,n}^2)$,
where the subscript $n$ indicates their dependency on the training sample size. Without loss of generality, we assume that for a given dataset $D_n=(\mathbb{Y},\mathbb{X})$, where $\mathbb{Y} \in \mathbb{R}^{n\times d_{h+1}}$ and $\mathbb{X} \in \mathbb{R}^{n\times p}$, 
the true model is a sparse StoNet with $\bsigma_n^2$ being known and satisfying Assumption \ref{ass:1}-(v).
\textcolor{black}{
\begin{remark} \label{remark:true_model}
 Since the StoNet is asymptotically equivalent to the DNN, as shown in Lemma \ref{lemma:2}, the true model we assume for the data is essentially a sparse DNN. We introduce StoNet primarily for its theoretical properties, which bridge sparse learning theory from linear models to DNNs as shown in Section \ref{consistency_section}.
 From this perspective, it becomes clear why we assume that $\bsigma_n^2$ is known, which ensures 
  necessary conditions are met and the existence of a sparse StoNet model with the desired accuracy of approximation to the underlying data-generating system. Notably, a sparse DNN can approximate many classes of functions, such as affine and 
$\alpha$-Hölder smooth functions, arbitrarily well as $n\to \infty$, as discussed in \citet{SunSLiang2021}. Given this approximation power and the asymptotic equivalence between two models, we assume that the true model is a sparse StoNet,
thereby avoiding the theoretical complexity in dealing with the DNN approximation errors.
 \end{remark} 
  }

The IRO algorithm starts with an initial estimate of $\btheta$, denoted by $\hat{\btheta}_n^{(0)}$, and then iterates between the imputation and regularized optimization steps as shown in Algorithm \ref{IROforstonet} 
given in the Supplement. 
 The key to the IRO algorithm is to find a sparse estimator for the working 
 true parameter $\btheta_*^{(t)}$, as defined in equation (\ref{IROeq2}), that is uniformly consistent over all iterations. As suggested by  \cite{Liang2018missing}, such a  uniformly consistent sparse estimator can typically be obtained by minimizing an appropriately penalized loss function as defined in (\ref{IROeq1}).
  For the StoNet, 
 solving (\ref{IROeq1}) corresponds to solving \emph{a series of  
 linear regressions} by noting that the joint distribution $\pi(\bYmis,\bY|\bX,\btheta,\bsigma_n^2)$ can be decomposed in a Markov structure:
 \begin{equation} \label{eq:decomp}
 \small
 \pi(\bYmis,\bY|\bX,\btheta,\bsigma_n^2)
 =\pi(\bY|\bY_h,\btheta,\bsigma_n^2) 
  \pi(\bY_h|\bY_{h-1},\btheta,\bsigma_n^2) 
 \cdots \pi(\bY_1|\bX,\btheta,\bsigma_n^2),
 \end{equation}
 and, furthermore, the components of $\bY_i \in \mathbb{R}^{d_{i,n}}$ are mutually independent conditional on $\bY_{i-1}$.

\subsection{Consistency of Sparse StoNets} 
\label{consistency_section}

Suppose that the Lasso penalty \citep{Tibshirani1996} 
is used in (\ref{IROeq1}). In this section, we aim to show
that the resulting IRO estimator $\hat{\btheta}_n^{(t)}$ is consistent when both $n$ and $t$ are sufficiently large, and that 
the sparse StoNet structure can be consistently identified. 
The proofs for all theoretical results are deferred to the 
Supplement. 
It is important to note that the parameter estimation consistency for neural networks is subject to loss-invariant transformations as defined in Section \ref{stonetsection}.

A critical step in establishing the consistency of parameter estimation for sparse StoNets is to verify that the estimator obtained from \eqref{IROeq1} consistently estimates $\btheta_{*}^{(t)}$ \eqref{IROeq2}. First, we introduce the assumptions required for this, drawing upon established literature on linear models, such as \cite{Meinshausen2009LASSOTYPERO}.
We define the $m$-sparse minimal and maximal eigenvalues for a matrix $\Sigma$ as follows:
\[
\small
\phi_{\min}(m|\Sigma)  = \min_{\bbeta: \|\bbeta\|_{0} \leq m} \frac{\bbeta^T \Sigma\bbeta }{\bbeta^T \bbeta}, \quad   
\phi_{\max}(m|\Sigma)  = \max_{\bbeta: \|\bbeta\|_{0} \leq m} \frac{\bbeta^T \Sigma\bbeta }{\bbeta^T \bbeta}, 
\]
 which represent, respectively, the minimal and maximal eigenvalues of any $m\times m$-dimensional principal submatrix. 
Let $\bSigma_n \in \mathbb{R}^{p_n\times p_n}$ denote the covariance matrix of the input variables. Let $q_{l,k,n}^{(t)}$ denotes the size of the working true regression formed for neuron $k$ of layer $l$ at iteration $t$, as implied by the working true parameter $\btheta_*^{(t)}$. Then we need the following assumption:

\begin{assumption} \label{ass:3} 
(i) The input variable $\bX$ is bounded, and there exist a constants $0< \kappa_{\min} <\infty$ 
such that $\liminf_{n \to \infty} \phi_{\min}(\min\{n,p_n\}|\bSigma_n) \geq \kappa_{\min}$;
(ii) there exists a sparse exponent $s\in [0,1]$ such that $q_{l,k,n}^{(t)} \prec d_{l-1,n}^s$ 
for $1\leq k \leq d_{l,n}$,
$1\leq l\leq h_n+1$, and any iteration $t$, and set $(\sigma_{1,n}^2, \sigma_{2,n}^2,\ldots,\sigma_{h_n+1,n}^2)$  such that the following conditions hold:  $\frac{\kappa_{\min}^2}{\sigma_{1,n}^2} \succ \frac{h_n d_{1,n} p_n^s \log p_n}{n}$, and  $\frac{\sigma_{l-1,n}^4}{\sigma_{l,n}^2} \succ \frac{h_n d_{l,n} d_{l-1,n}^s \log d_{l-1,n}}{n}$ for  $l \in \{2,3,\ldots,h_n+1\}$;
(iii) the activation function $\psi(\cdot)$ is bounded.
\end{assumption}

Assumption \ref{ass:3}-(i) is a standard condition commonly used in high dimensional variable selection literature, as seen in works like 
\cite{jiang2007bayesian} and \cite{huang2008iterated}.
Assumption \ref{ass:3}-(ii) works with  Assumption \ref{ass:1}-(v) to constrain the range of 
$\sigma_{l,n}$'s. Notably, such a uniform sparse exponent $s$ always exists and, in the worst case, can be set to 1. 
Assumption \ref{ass:3}-(iii) is primarily a technical condition. 
Since $\sigma_{l,n}^2$'s are typically set to very small values, 
it is easy to confine the random errors  $\be_i$'s to a compact space with high probability. Consequently, an unbounded activation function such as {\it ReLU} can still be used in the StoNet, though the corresponding theoretical results need to be slightly modified to hold with high probability. 

Under Assumptions \ref{ass:1} and  \ref{ass:3}, we prove the following lemma, which bounds the $m$-sparse minimal and maximal eigenvalues of the covariance matrix of the latent variables at each hidden layer.

 \begin{lemma} \label{lem:eigen_hidden_layer} For any $l \in \{1,2,\ldots,h\}$, we let $\bSigma_l^{(t)}$ denote the sample covariance matrix of the covariates of the linear regressions formed for each neuron of layer $l+1$ at iteration $t$.
If Assumption \ref{ass:1} and Assumption \ref{ass:3} hold, then there exist constants $c>0$ and  $0<\kappa_{\max,l} < \infty$ such that for any iteration $t$,  
\[
\small
\phi_{\min}( \min\{n,d_{l,n}\}|\bSigma_l^{(t)}) \geq c \sigma_{l,n}^2, \quad 
\phi_{\max}(\min\{n,d_{l,n}\}|\bSigma_l^{(t)}) 
\leq \kappa_{\max,l}.
\]
\end{lemma}

\begin{remark} \textcolor{black}{In practice, the ReLU activation function can also be used  as justified below. 
At each iteration $t$, if a hidden neuron belongs to the true neuron set $\bS_t$ as determined by $\btheta_*^{(t)}$, then $\Psi(\tilde{\bY}_l)$ cannot be constantly 0 
over all $n$ samples. Therefore, it is reasonable to assume that there exists a 
threshold $q_{\rm min} \in (0,1)$ such that $\mathbb{E}[\nabla_{\tilde{\bY}_{l}^{(t,i)}}\Psi(\tilde{\bY}_{l}^{(t,i)}) \circ \nabla_{\tilde{\bY}_{l}^{(t,i)}}
\Psi(\tilde{\bY}_{l}^{(t,i)})] \geq q_{\rm min}$ for any true neuron in all iterations, where $t$ and $i$ index the iteration and neuron, respectively. 
Under this assumption, we would at least have 
\[
\small
\phi_{\rm min}(|\bs_l^{(t)}| | \Sigma_l) \geq \kappa_{\min,l}:=\sigma_{l,n}^2 q_{\rm min},  \quad l=1,2,\ldots,h; \ \ t=1,2,\ldots, T, 
\]
where $\bs_l^{(t)} \subset \bS^{(t)}$ denotes the set of true neurons at layer $l$. 
As implied by the proof of Lemma \ref{lem:eigen_hidden_layer}, the maximal eigenvalue 
$\phi_{\max}(\min\{n,d_{l,n}\}|\bSigma_l^{(t)}) 
\leq \kappa_{\max,l}$ still holds for the ReLU activation function.
In consequence, Lemma \ref{lem:eigen_hidden_layer} and the followed Theorem \ref{thm:stonet-IRO} can still hold with the ReLU activation function. 
 }
\end{remark}

Regarding the setting of regularization parameters,
we have the following assumption which directly follows from the theory developed by \cite{Meinshausen2009LASSOTYPERO} for linear regression and \cite{huang2008iterated} for logistic regression.  

\begin{assumption} \label{ass:4} The Lasso penalty is used for the StoNet. At each iteration $t$, (i) set the regularization parameter $\lambda_{l,n}^{(t)}\asymp \sigma_{l,n} (n \log d_{l-1,n})^{1/2}$ for each linear regression layer $l$; and (ii) set the regularization parameter $\lambda_{l,n}^{(t)}\asymp (n^{2+\varepsilon} \log d_{l-1,n})^{1/3}$  for some $\varepsilon \in (0,1)$ for each logistic regression layer. \end{assumption}

Further, let's consider the mapping $M(\btheta)$ as defined in the 
EM-update (\ref{IROeq2}), i.e., 
\[
\small
M(\btheta)=\arg\max_{\btheta'}  \mbE_{\btheta} \log\pi(\bY,\bYmis |\bX,\btheta',\bsigma_n^2),
\]
\textcolor{black}{where the expectation is taken with respect to the conditional distribution $\pi(\bY|\bX)\pi(\bYmis|\bY,\bX,\btheta)$, and
    $\pi(\bYmis|\bY,\bX,\btheta)$ is defined by the StoNet.}

\begin{assumption} \label{ass:5}  The mapping $M(\btheta)$ is differentiable. Let $\rho_{\max}(\btheta)$ be the largest singular value of $\partial M(\btheta)/\partial \btheta$. There
exists a number $\rho^* <1$ such that  $\rho_{\max}(\btheta) \leq \rho^*$ for all 
$\btheta \in \Theta_n$ for sufficiently large $n$ and almost every $D_n$ observation sequence.
\end{assumption}

As discussed in \cite{Nielsen2000}, the differentiability 
of $M(\btheta)$ ensures the EM-update to be a contraction.
A recursive application of the mapping, i.e., setting
$\btheta_n^{(t+1)}=\btheta_*^{(t+1)}=M(\btheta_n^{(t)})$, leads to a monotone increase of the target expectation $\mbE_{\btheta_n^{(t)}} \log \pi(\bY,\bYmis |\bX,\btheta_n^{(t+1)},\bsigma_n^2)$ along with the convergence of $\btheta_n^{(t)}$ to a fixed point, a property well-established for the EM algorithm \citep{Dempster1977,Wu1983}. 
Moreover, the continuity of $M(\btheta)$ ensures that $\rho_{\max}(\btheta)<1$ holds in a neighborhood of the fixed point. 
Suppose the mapping possesses multiple fixed points, 
we expect that each of them corresponds to an optimal solution equivalent to $\btheta^*$, 
up to loss-invariant transformations. 
Thus, each fixed point can be mathematically regarded as unique. This leads to the following theorem.

\begin{theorem} \label{thm:stonet-IRO} Suppose that the Lasso penalty is imposed on $\btheta_n$, and Assumptions \ref{ass:1}-\ref{ass:4} hold.
\begin{itemize}  
\item[(i)] There exist some constants $c_1>0$, $c_2>0$, and $c_3>0$ such that $\mathbb{E}(\|\hat{\btheta}_n^{(t)} - \btheta_*^{(t)} \|_2^2) \prec r_n=o(1)$ holds uniformly for all iterations with 
\begin{equation}
r_n=c_1 \frac{\sigma_{1,n}^2}{\kappa_{\min}^2}  d_{1,n} p_n^s \frac{\log p_n}{n} + c_2 \sum_{l=2}^{h+1} \frac{\sigma_{l,n}^2}{\sigma_{l-1,n}^4} d_{l,n} d_{l-1,n}^s \frac{\log d_{l-1,n}}{n},
\end{equation}
for the StoNet with a linear regression output layer, and 
\begin{equation}
\begin{split}
r_n = & c_1 \frac{\sigma_{1,n}^2}{\kappa_{\min}^2}  d_{1,n} p_n^s \frac{\log p_n}{n} + c_2
\sum_{l=2}^{h} \frac{\sigma_{l,n}^2}{\sigma_{l-1,n}^4} d_{l,n} d_{l-1,n}^s \frac{\log d_{l-1,n}}{n} \\
 & \quad + \frac{c_3}{\sigma_{h,n}^4} d_{h+1,n} d_{h,n}^s \frac{(\log d_{h,n})^{2/3}}{n^{2(1-\varepsilon)/3}},
\end{split}  
\end{equation}
for the StoNet with the logistic regression output layer.

\item[(ii)] Furthermore, if Assumption \ref{ass:5} holds, then $\|\hat{\btheta}_n^{(t)} - \btheta^* \| \stackrel{p}{\to} 0$ for sufficiently large $n$ and sufficiently large $t$ and almost every training dataset $D_n$.  
\end{itemize}
\end{theorem}

To establish the structure selection consistency for the sparse
StoNet, we need the following $\theta$-min condition:

\begin{assumption} \label{ass:6}  ($\theta$-min condition) 
$\min_{k \in \bgamma^*} |\theta_k^*| \succ 
\sqrt{r_n}$, where $\bgamma^{*} = \{k: \theta_k^{*} \neq 0 \}$ is the set of indexes of non-zero elements of $\btheta^*$ and $\theta_k^*$ denotes the $k$-th component of $\btheta^*$.
\end{assumption} 

Assumption \ref{ass:6} is essentially an identifiability condition, which ensures   non-zero elements of $\btheta^*$ can be distinguished from $0$. This is a typical condition in high-dimensional variable selection, see e.g., \cite{zhao2006model}.  

\begin{corollary}\label{cor:stonet-IRO} 
Suppose that the conditions of 
 Theorem \ref{thm:stonet-IRO} hold. If Assumption \ref{ass:6} also holds and the connections are selected by setting  
$\widehat{\bgamma}_{n}^{(t)}:=\{i: 
|\hat{\theta}_{i,n}^{(t)}| > c \sqrt{r_n} \}$ 
for some constant $c$, where $\hat{\theta}_{i,n}^{(t)}$ denotes the $i$-th component of $\hat{\btheta}_n^{(t)}$, then the selected model 
$\widehat{\bgamma}_n^{(t)}$ is a consistent estimator of the true  model \textcolor{black}{$\bgamma^* := \{i: |\theta^*_i| \neq 0\}$}. That is, $P(\widehat{\bgamma}_n^{(t)} = \bgamma^*)\rightarrow 1$ as $n\to \infty$ and $t \to \infty$.
\end{corollary}
 
 As suggested by Theorem \ref{thm:stonet-IRO} and Corollary \ref{cor:stonet-IRO}, we have provided a constructive proof for  parameter estimation and structure selection consistency for sparse StoNets, based on the sparse learning theory of linear models. 
 Moreover, due to the 
 asymptotic equivalence between the StoNet 
 and DNN models (see Lemma \ref{lemma:2}), the consistency results in Theorem \ref{thm:stonet-IRO} and Corollary \ref{cor:stonet-IRO} also apply to DNNs, although this is not the focus of the present paper. Nonetheless, Theorem \ref{thm:stonet-IRO} and Corollary \ref{cor:stonet-IRO} show how the StoNet can facilitate the transfer of theory and methods from linear models to DNNs.

 It is important to note that 
 Theorem \ref{thm:stonet-IRO} provides a theoretical guarantee for the validity of  
 post-StoNet modeling in uncertainty quantification. Without the consistency of parameter estimation,
 as shown in Section 
 \ref{sect:non-consistency} of the Supplement, the resulting confidence interval can be dishonest. 

\begin{remark} \label{rem:SGMCMC}
While the IRO algorithm facilitates the transfer of statistical theory from linear models to DNNs, it is less scalable with respect to 
big data due to its requirement of full data computation at each iteration.
 To address this issue, we can train the sparse StoNet using an adaptive stochastic gradient MCMC algorithm, which is 
designed to find the 
estimator 
$\hat{\btheta}_n^*=\arg\max_{\btheta} \{\pi(\bY|\bX,\btheta,\bsigma^2) P_{\lambda_n}(\btheta)\}$ by solving the equation:  
\begin{equation} \label{rooteq}
\small
\int \nabla_{\btheta} [\log \pi(\bY,\bYmis|\bX,\btheta,\bsigma^2)+\log P_{\lambda_n}(\btheta)] \pi(\bYmis|\bY,\bX,\btheta,\bsigma^2) d\bYmis=0.
\end{equation} 
We note that solving (\ref{rooteq}) is equivalent to solving   
$\nabla_{\btheta}[ \log \pi(\bY|\bX,\btheta, \bsigma^2)+$ $\log p_{\lambda_n}(\btheta)]=0$
 by Fisher's identity. 
This algorithm is scalable with respect to big data by making use of mini-batch samples at each iteration. 
Since the algorithm is a slight extension of the adaptive stochastic gradient Hamiltonian Monte Carlo algorithm 
in \cite{LiangSLiang2022}, we leave it to the Supplement (see Algorithm \ref{ASGHMC}). 
Following from the convergence theories  of the IRO and  adaptive stochastic gradient MCMC algorithms,
it is straightforward to show that Algorithm \ref{ASGHMC} also  
yields a consistent estimate of $\btheta^*$ as well as a consistent 
recovery for the sparse StoNet structure $\bgamma^*$. In practice, Algorithm \ref{ASGHMC} can also be used as an efficient tool for generating good initial values for the IRO algorithm. 
\end{remark}

\subsection{Prediction Intervals of Sparse StoNets}  
\label{sect:CI}

The hierarchical structure of the StoNet enables us to quantify the uncertainty of the latent variables at each layer recursively using Eve's law. 
Consider a StoNet model trained by the IRO algorithm, 
let $\bz$ denote a test point at which the prediction uncertainty is to be quantified.
Let $\bZ_i^{(t)}$ denote the latent variable at layer $i$, 
which corresponds to the test point $\bz$ and is imputed based on the parameter $\hat{\btheta}^{(t)}$ at iteration $t$ of 
the IRO algorithm. Let $\bmu_i^{(t)}$ and $\bSigma_i^{(t)}$ denote, respectively, the mean and covariance matrix of $\bZ_i^{(t)}$. By Eve's law, for any layer $i \in \{2, 3, \dots, h+1\}$,  we have $\bSigma_{i}^{(t)}=\mbE (\Var(\bZ_{i}^{(t)}|\bZ_{i-1}^{(t)})) + \Var(\mbE(\bZ_{i}^{(t)}|\bZ_{i-1}^{(t)}))$. 
As detailed in Section \ref{sect:CI-APP} (of the Supplement), the 
estimator $\widehat{\bSigma}_{i}^{(t)}$, given in (\ref{varest}), can be 
derived. This leads to the following procedure for prediction interval construction.

Let $\mu(\bz,\hat{\btheta})$ denote the prediction  of a StoNet with weights 
 $\hat{\btheta}$ at the test point $\bz$. 
Note that the StoNet (\ref{eq:stonet}) has the same prediction function as the DNN (\ref{eq:DNN}), i.e., the random noise added to the latent variables is set to 0 in forward prediction.  Suppose that a set of StoNet estimates, $\mS=\{\hat{\btheta}^{(1)},\hat{\btheta}^{(2)}, \ldots,\hat{\btheta}^{(m)} \}$, 
has been collected after convergence of the IRO algorithm. 
By the Wald method, the 95\% prediction interval of 
$\mu_j(\bz,\btheta^*)$, the $j$-th component of $\mu(\bz,\btheta^*)$, 
can  be constructed in  Algorithm \ref{alg:CI}:

\begin{algorithm}[!ht]
\caption{\textcolor{black}{Prediction Interval Construction with StoNet}}
\label{alg:CI}
 \KwInput{A set of StoNet estimates: $\mS=\{\hat{\btheta}^{(1)},\hat{\btheta}^{(2)}, \ldots,\hat{\btheta}^{(m)} \}$ and a test point $\bz$.} 
\For{$t=1,2,\ldots,m$}{
$\bullet$ Calculate the covariance of latent variables: $\widehat{\bSigma}_i^{(t)}$, $i=1,2,\ldots,h+1$.\\
\For{$j=1,2,\ldots,d_{h+1}$}{
 $\bullet$  Calculate the variance of training error: $\hat{\varsigma}_{h+1,j}^{2(t)}= \frac{1}{n} \sum_{k=1}^n \left(\mu_j\bigl(\bx^{(k)},\hat{\btheta}^{(t)}\bigr)-y^{(k)}_j\right)^2$, 
where $\bigl(\bx^{(k)}, \by^{(k)}\bigr)$ denotes the $k$-th training sample, and 
$y_j^{(k)}$ denotes the $j$-th component of $\by^{(k)}$.

 $\bullet$ Calculate the 95\% prediction interval:  
\begin{equation} \label{intervaleq}
\small 
\left(\mu_j\bigl(\bz,\hat{\btheta}^{(t)}\bigr) - 1.96\sqrt{\widehat{\bSigma}_{h+1,j}^{(t)} + \hat{\varsigma}_{h+1,j}^{2(t)}}, 
 \mu_j\bigl(\bz,\hat{\btheta}^{(t)}\bigr) + 1.96\sqrt{\widehat{\bSigma}_{h+1,j}^{(t)} + \hat{\varsigma}_{h+1,j}^{2(t)}}\right),
\end{equation}
where $\widehat{\bSigma}_{h+1,j}^{(t)}$ denotes the $(j,j)$-th diagonal element of $\widehat{\bSigma}_{h+1}^{(t)}$.
} } 
\KwOutput{Output 95\% prediction intervals for $\mu_j(\bz,\btheta^*)$, $j \in\{1,2,\ldots,d_{h+1}\}$, by averaging respective $m$ intervals.}  
\end{algorithm}

The honesty of the resulting prediction interval follows from the consistency of $\widehat{\btheta}^{(t)}$ as established in Theorem \ref{thm:stonet-IRO}.
\textcolor{black}{
\begin{corollary} \label{cor2:CI}
 Suppose that a sparse StoNet estimator $\hat{\btheta}$ is consistent. Then, for 
 any test point $\bz$ distributed as  validation samples and any $j \in \{1,2,\ldots, d_{h+1}\}$,  
 \[
 \small
 P\left(\mu_j(\bz,\btheta^*) \in \Big\{\mu_j(\bz,\hat{\btheta})\pm c_{\alpha} 
 \sqrt{ \widehat{\bSigma}_{h+1,j}+ \varsigma_{h+1,j}^{2}} \Big\} \right) 
 -(1-\alpha) \to 0, 
 \]
 as $n \to \infty$, where $c_{\alpha}=\Phi^{-1}(1-\alpha/2)$ denotes the 
 $(1-\alpha/2)$-quantile of the standard Gaussian distribution. 
\end{corollary}
}
Note that the dependence of $\hat{\btheta}$, $\widehat{\bSigma}_{h+1,j}$, 
and $\varsigma_{h+1,j}$ on the sample size $n$ is implicit, and this 
relationship is depressed in Corollary \ref{cor2:CI} for notational simplicity.
Algorithm \ref{alg:CI} can be easily extended to the StoNet with a logistic regression output layer via the Wald/endpoint transformation. By Remark \ref{rem:SGMCMC}, both the IRO and adaptive stochastic gradient MCMC algorithms asymptotically converge to the same solution, 
the above procedure can also be applied to the StoNet trained by Algorithm \ref{ASGHMC}.

\subsection{Uncertainty Quantification of Large-Scale DNNs} 
\label{post_stonet_section}

In real applications, the use of large-scale DNNs, such as residual networks \citep{he2016deep} and transformer \citep{dosovitskiy2020image}, 
has been a common practice. Training  large-scale models from scratch can be expensive. Moreover, these large-scale models are often miscalibrated \citep{guo2017calibration}. In the spirit of improving the model without significantly modifying the standard training procedure, we propose a post-StoNet approach, which applies sparse StoNet as a post-processing tool to quantify the prediction uncertainty of trained DNN models. The proposed approach works in Algorithm \ref{alg:post-stonet}:

\begin{algorithm} 
\caption{\textcolor{black}{The post-StoNet approach}}
\label{alg:post-stonet}
 \KwInput{A pre-trained DNN model, a validation dataset, and a test dataset.} 

 {\it $\bullet$ Feature transformation:}  Transform the input variables of the validation and test samples by extracting the output from the last hidden layer of the pre-trained DNN model.

 {\it $\bullet$ Sparse StoNet modeling:}  
 For the validation dataset, model the relationship between the response variable and the transformed features using a simple sparse StoNet (e.g., with one hidden layer only). Train the model using either Algorithm \ref{ASGHMC} or Algorithm \ref{IROforstonet}, and collect a set of 
 sparse StoNet estimates. 
 
{\it $\bullet$ Prediction interval construction:} 
For each data point in the test dataset, use Algorithm \ref{alg:CI} to construct a prediction interval based on the transformed features. 

\KwOutput{A prediction interval for each data point in the test dataset.} 
\end{algorithm}

The rationale underlying Algorithm \ref{alg:post-stonet} can be justified as follows: As shown in \cite{LiangSLiang2022}, 
the last-hidden-layer's output of the StoNet serves as 
 a nonlinear sufficient dimension reduction of the input data. Building upon the asymptotic equivalence between 
the StoNet and DNN, 
the transformed data from
a well-trained DNN approximates a sufficient dimension reduction of the input data. The DNN typically gives a simple linear relationship between   transformed data and response, but this  relationship may no longer hold for validation data due to the possible over-fitting issue. Therefore, we remodel it  using a simple sparse StoNet, enabling the prediction uncertainty to be correctly quantified.

\vspace{-0.2in}
\section{An Illustrative Example} \label{simulationsection}

 This example serves as a validation of our consistency theory established for sparse StoNets. Consider two neural network models:
\begin{align} 
\small
y  & =  2 \tanh(2 x_1 - x_2) + 2 \tanh(x_3 - 2 x_4)  - \tanh(2 x_5)+ 0 x_6 + \dots 0 x_{20}+\epsilon, \label{Exeq1} \\
 y  & =  \tanh(2 \tanh (2x_1 - x_2)) + 2 \tanh(2 \tanh(x_3- 2 x_4) - \tanh(2 x_5)) + 0 x_6 \nonumber \\ 
 & \quad + \dots 0 x_{20}+\epsilon, \label{Exeq2}
\end{align}
where $\epsilon \sim N(0,1)$, $\bx=(x_1,x_2,\ldots,x_{20})$, 
$x_i \sim N(0,1)$ for $i=1,2,\ldots, 20$, and $x_i$'s are correlated with a mutual correlation coefficient of 0.5.  Equations (\ref{Exeq1}) and (\ref{Exeq2}) represent a one-hidden-layer neural network and a two-hidden-layer neural network,  respectively. For both models, the variables $x_1, x_2, \ldots, x_5$ are true and the others are false. The strong mutual correlation makes 
the true variables hard to identify. 
From each model, we simulated 1000 samples, 500 samples for training and 500 samples for test. 
 We use this example to assess the performance of the StoNet in nonlinear variable selection and prediction uncertainty quantification, and to examine the effect of $\bsigma^2=(\sigma_1^2,\ldots,\sigma_{h+1}^2)$ on the performance of the StoNet as well. 
 
We modeled the data from model (\ref{Exeq1}) by a StoNet with structure 20-500-1, and that from model (\ref{Exeq2}) by a StoNet with structure 20-500-100-1.  We trained the StoNets by Algorithm  \ref{ASGHMC} with different parameter settings as given in Section \ref{sect:settingsimu}. 
For convenience, we call these settings 
(i), (ii), and (iii), respectively. 
Under each setting, Algorithm \ref{ASGHMC} was run for 2000 epochs with 
the Lasso penalty $P_{\lambda}(\btheta) = \lambda \|\btheta\|_1$.  Various values of $\lambda$ have been tried for the example as described below. 

\begin{figure}[!h]
    \centering
    \begin{tabular}{cc}
    (a) one-hidden-layer StoNet & (b) two-hidden-layer StoNet \\
       \includegraphics[width=2.5in,height=1.8in]{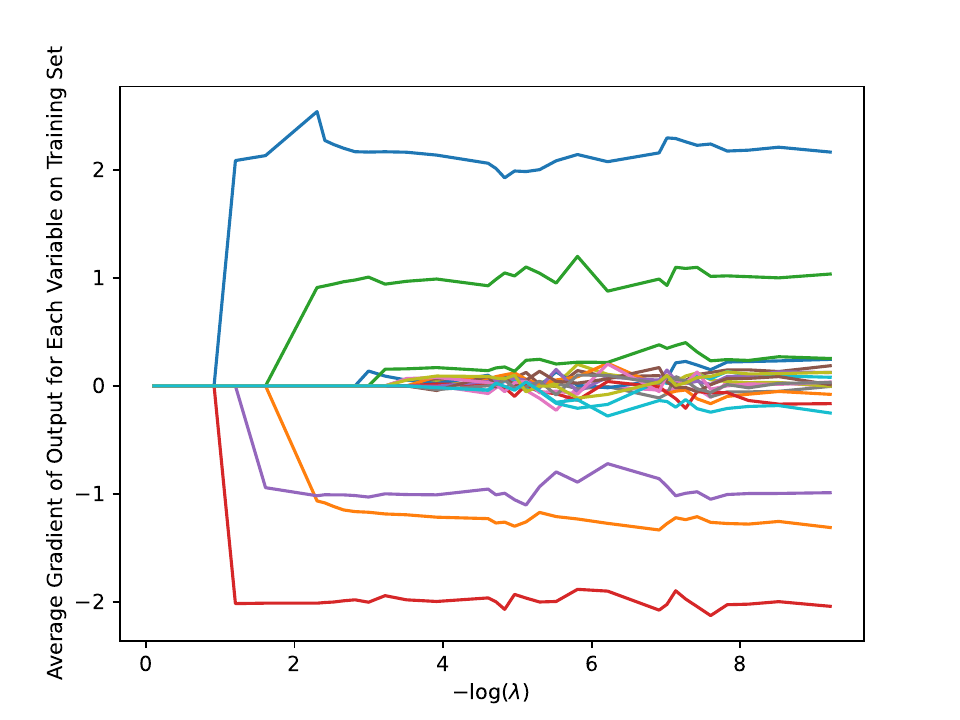}  & 
     \includegraphics[width=2.5in,height=1.8in]{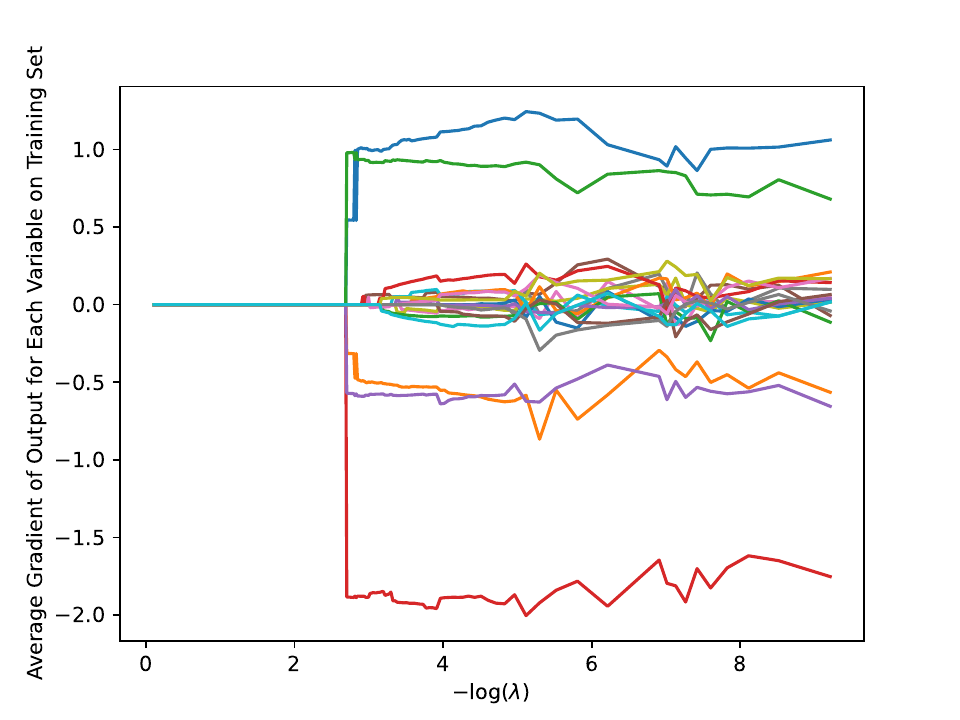} 
    \end{tabular}
    \caption{Variable selection paths by the StoNet under setting (ii)  for the model (\ref{Exeq1}) (left plot)  and
    the model (\ref{Exeq2}) (right plot),
     where $y$-axis is the average output gradient $\frac{1}{n}\sum_{k=1}^n \frac{\partial \hat{\mu}(\bx)}{\partial x_i}|_{\bx^{(k)}}$ calculated over the training data, and $x$-axis is $-\log(\lambda)$.
    }
    \label{Lasso_selection_path}
\end{figure}

To examine the effect of $\lambda$, we mimic the regularization path for LASSO and measure the importance of each variable by the average output gradient $\frac{1}{n}\sum_{k=1}^n \frac{\partial \hat{\mu}(\bx)}{\partial x_i}|_{\bx^{(k)}}$ calculated over training samples, where $\hat{\mu}(\bx)$ denotes the forward prediction function of the StoNet and $\bx^{(k)}$ denotes the $k$th-observation of the training set. Using the partial derivative to evaluate the dependence of a function on a particular variable has been proposed 
by \cite{RosascoLorenzo2013NonparametricSA}, and employed in \cite{Zheng2020LearningSN} for sparse graphical modeling. Figure \ref{Lasso_selection_path} shows the regularization paths of the StoNet for the models (\ref{Exeq1}) and (\ref{Exeq2}).    
 Further, we examined the paths of each variable and found that for both models, the true variables $x_1, x_2,\ldots, x_5$ can be correctly identified by the StoNet with an appropriate value of $\lambda$.

\begin{figure}[!ht] 
    \centering
    \begin{tabular}{cc} 
   (a) & (b) \\ 
    \includegraphics[width=2.5in, height=1.5in]{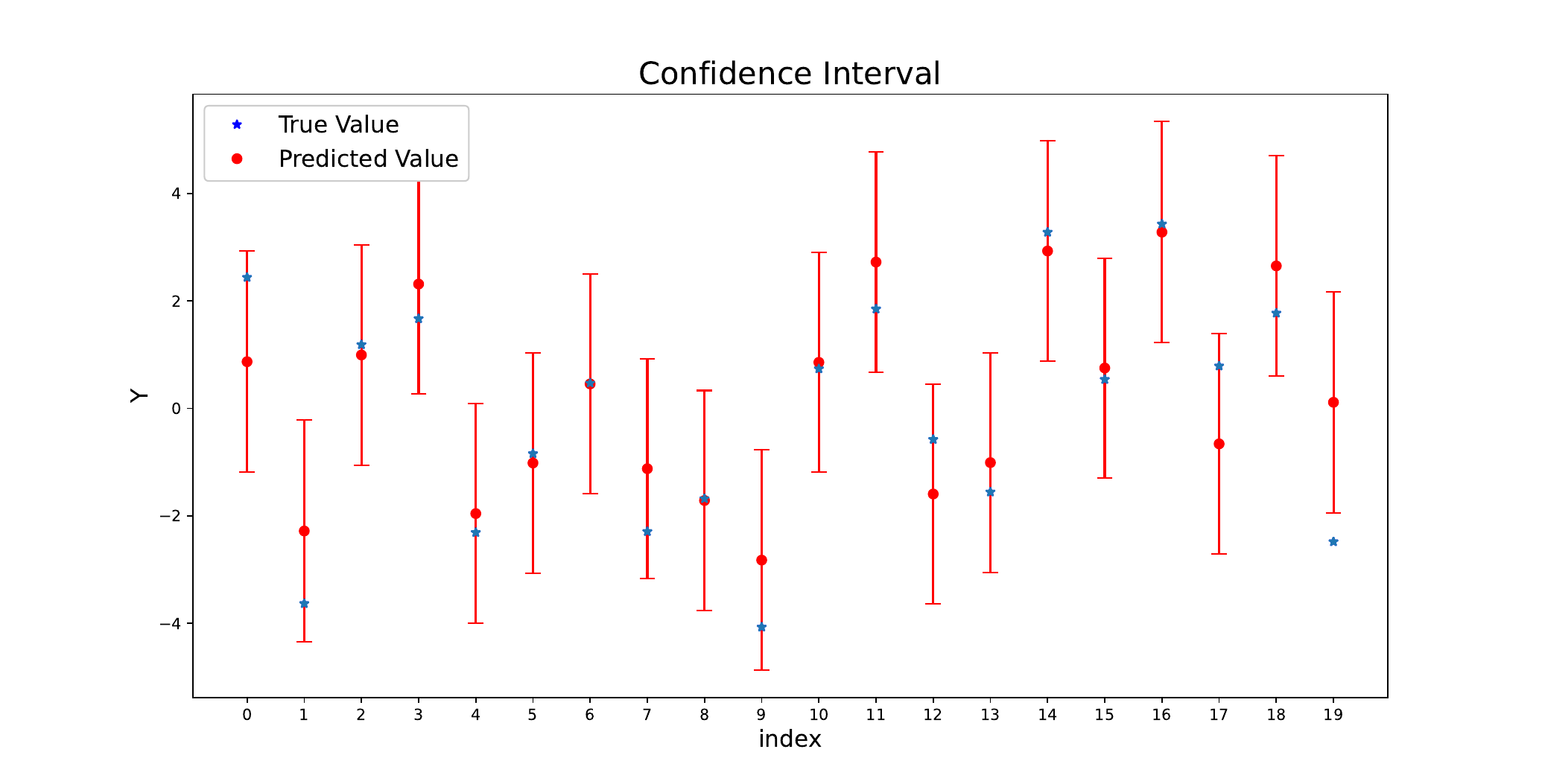} & 
    \includegraphics[width=2.5in, height=1.5in]{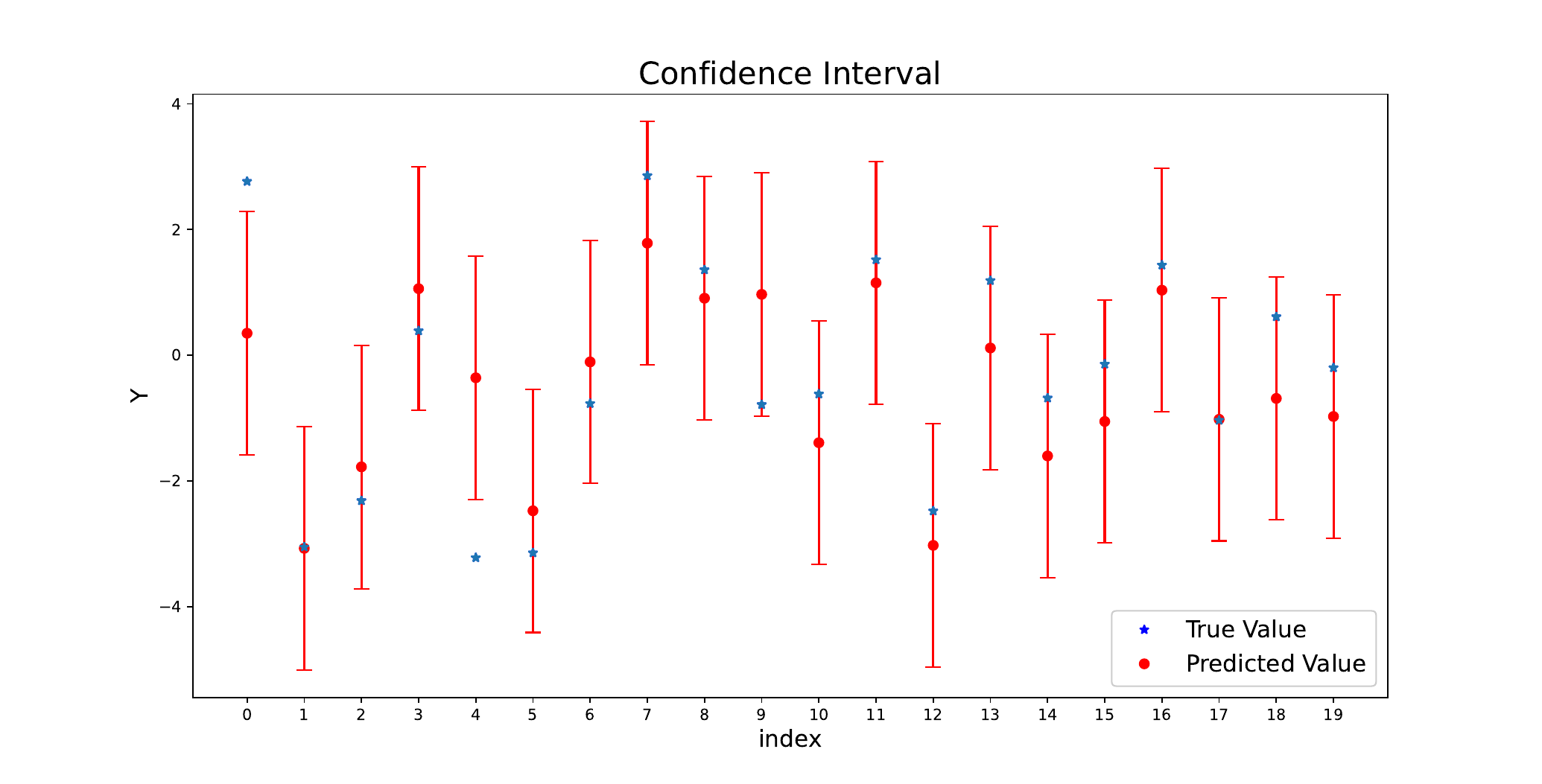}
    \end{tabular}
    \caption{Prediction intervals produced by (a) one-hidden-layer StoNet and (b) two-hidden-layer StoNet at 20 test points, where the StoNets were trained under setting (ii).}
    \label{stonet_CI}
    \vspace{-0.3in}
\end{figure}

Next, we examined the performance of the StoNet in prediction uncertainty quantification. 
We generated 100 training datasets, each consisting of 500 \textcolor{black}{training}  samples, from each of 
the models (\ref{Exeq1}) and (\ref{Exeq2}). 
For each training dataset, a StoNet was trained as described above, 
and a prediction interval was constructed for each sample point of the test dataset 
with the StoNet estimate obtained at the last iteration of the run.
Table \ref{CItab} summarizes the prediction intervals obtained at $500$ test points for respective models. 
Under all three parameter settings, the StoNet models provide coverage rates close to the target level, which demonstrates the stable performance of the StoNet model.
As expected, the StoNet produces better coverage rates with smaller values of $\bsigma^2$, 
since the data were simulated from DNN models. Figure \ref{stonet_CI} shows the prediction intervals produced by the StoNet at some test points, indicating the excellence of the StoNet in prediction uncertainty quantification.  

\begin{table}[!ht]
\caption{Coverage rates of 95\% prediction intervals for 500 test samples simulated from the models (\ref{Exeq1}) and (\ref{Exeq2}), with the corresponding standard deviations given in parentheses.} 
\centering
\vspace{2mm}
\begin{tabular}{ccccc} \toprule
Model  &   \textcolor{black}{Setting (i)}&   \textcolor{black}{Setting (ii)} &   \textcolor{black}{Setting (iii)} \\ 
\midrule
{Model (\ref{Exeq1})} & 94.766\% (2.157\%) & 94.496\% (2.162\%)& 94.310\% (2.197\%) \\ 
\midrule
{Model (\ref{Exeq2})} & 94.642\% (2.189\%) & 94.396\% (2.256\%) & 94.300\% (2.290\%) \\ 
\bottomrule
\end{tabular}
\label{CItab}
\end{table}

\section{Numerical Experiments} \label{sect:examples}

This section demonstrates that the post-StoNet approach improves model calibration and provides shorter prediction intervals 
compared to the conformal method.

\subsection{Classification Problems} 
We conducted experiments on the CIFAR100 dataset. Following the setting of post-calibration methods in \cite{guo2017calibration}, we split the training data into a training set of 45000 images and a hold-out validation set of 5000 images. We modeled the data using DenseNet40 \citep{huang2017densely},
ResNet110 \citep{he2016deep} and WideResNet-28-10 \citep{zagoruyko2016wide}, which consist of 1.7M, 176K, and 36M 
 parameters, respectively. 
Refer to Section \ref{parametersection} for hyperparameter settings
for the StoNet. 
For comparison, we considered post-hoc calibration methods, including temperature scaling \citep{guo2017calibration}, matrix scaling \citep{guo2017calibration}, 
and some regularization-based calibration methods, including models trained with Focal loss \citep[][Focal]{mukhoti2020calibrating} and maximum mean calibration error penalty \citep[][MMCE]{kumar2018trainable}.
We repeated the experiment 10 times and reported  
the mean and standard deviation of prediction accuracy (ACC), negative log-likelihood loss (NLL), expected calibration error \citep[][ECE]{guo2017calibration} and class-wise expected calibration error \citep[][CECE]{kull2019beyond} in Table \ref{post_stonet_result_100}.
The results show that post-StoNet   significantly improves model calibration, especially in terms of ECE and CECE. \textcolor{black}{
For a thorough comparison, we also implemented a post-Linear method for the transformed data. This method follows the same procedure as post-StoNet, except that the sparse StoNet is replaced with a sparse multi-class logistic regression model with LASSO penalty. The regularization parameter for the LASSO penalty is selected via 5-fold cross-validation. The comparison shows that post-StoNet significantly outperforms the post-Linear method, highlighting the effectiveness of the StoNet in post-transformation data modeling.}

We have also conducted the same experiments 
to CIFAR10 with the results  
presented in Table \ref{post_stonet_cifar10} (in the Supplement). 
The comparison shows again that post-StoNet 
significantly improves model calibration, compared to existing methods.

\begin{table}[!ht]
\caption{Calibration results for CIFAR100 data, where the numbers in the parentheses represent the standard deviations of 
respective measures.}
\centering
\begin{adjustbox}{width=1.0\textwidth}
{\begin{tabular}{ccccccc}
\toprule
    Network & Size & Method & ACC(\%) & NLL & ECE(\%) & CECE(\%) \\
    \midrule
  \multirow{6}{*}{DenseNet40}    &  \multirow{7}{0.75cm}{176K} 
        & No Calibration & {\bf 68.32   (0.22)} & 1.5538 (0.0191) & 17.01   (0.25) & 0.404   (0.004)\\
      & & Temp. Scaling & {\bf 68.32   (0.22) } & {\bf 1.1615 (0.0104)} & 4.49   (0.49) & 0.220   (0.006)\\
      & & Matrix Scaling & 53.06   (0.83) & 5.2270 (0.4786) & 37.68   (1.20) & 0.818   (0.023)\\
      & & Focal & 67.41   (0.17) & 1.2098 (0.0105) & 7.12   (0.29) & 0.261   (0.002  )\\
      & & MMCE & 67.77   (0.42) & 1.1691 (0.0105) & 4.37   (0.30) & 0.217   (0.004)\\
      & & \textcolor{black}{Post-Linear} & 60.57 (0.25) & 1.4351 (0.0134) & 5.38 (0.40) & 0.266 (0.006) \\
      & & Post-StoNet & 68.13   (0.30) & 1.1972 (0.0104) & {\bf 1.64   (0.26)} & {\bf 0.188   (0.003)}\\
   \midrule
     \multirow{6}{*}{ResNet 110}    &  \multirow{7}{0.75cm}{1.7M} 
        & No Calibration & 71.68   (0.88) & 1.4923 (0.0434) & 16.67   (0.39) & 0.390   (0.008)\\
      & & Temp. Scaling & 71.68   (0.88) & {\bf 1.0488 (0.0339) } & 4.37   (0.35) & 0.209   (0.004)\\
      & & Matrix Scaling & 54.46   (3.12) & 5.2040 (0.5168) & 36.98   (2.51) & 0.800   (0.055)\\
      & & Focal & 70.81   (1.46) & 1.0738 (0.0581) & 6.04   (0.66) & 0.236   (0.012)\\
      & & MMCE & 68.14   (2.34) & 1.1848 (0.0865) & 4.76   (0.48) & 0.231   (0.009)\\
      & & \textcolor{black}{Post-Linear} & 57.29 (1.66) & 1.6490 (0.1602) & 9.35 (6.63) & 0.328 (0.109)\\
      & & Post-StoNet & {\bf 72.03   (1.26) }& 1.0520 (0.0374) & {\bf 1.11   (0.31) }  & {\bf 0.173   (0.002) }\\
    \midrule
  \multirow{6}{*}{WideResNet-28-10}  & \multirow{7}{0.75cm}{36M} 
       & No Calibration & 77.48   (0.88)  & 0.9307 (0.0361) & 9.30   (0.46) & 0.249   (0.009)\\
     & & Temp. Scaling & 77.48   (0.88) & 0.8662 (0.0349) & 4.93   (0.29) & 0.200   (0.008)\\
     & & Matrix Scaling & 68.99   (0.44) & 3.6928 (0.2287) & 25.07   (0.58) & 0.546   (0.011)\\
     & & Focal & 78.18   (0.16) & 0.9074 (0.0136) & 8.64   (0.22) & 0.239   (0.005)\\
     & & MMCE & 78.23   (0.26) & 0.8452 (0.0093) & 4.43   (0.23) & 0.191   (0.006)\\
     & & \textcolor{black}{Post-Linear} & 73.56 (0.44) & 2.3080 (0.0655) & 19.62 (0.51) & 0.440 (0.008) \\
     & & Post-StoNet &  {\bf 79.28   (0.62) } & {\bf 0.7580 (0.0202) } & {\bf 1.67   (0.29) }& {\bf 0.141   (0.004) }\\
\bottomrule
\end{tabular}}
\label{post_stonet_result_100}
\end{adjustbox}
\end{table}

\subsection{Regression Problems} 
 We used 10 datasets, with the sample size ranging from hundreds to hundreds of thousands, from the UCI repository. 
For each dataset, we randomly split the data into 40\% as the training set, 40\% as the validation set, and 20\% as the test set. We first trained a DNN model on the training set and then applied the post-StoNet approach on the validation  set to generate 90\% prediction intervals. We repeated the experiments 10 times and reported the mean and standard deviation of the coverage rates and prediction interval lengths in 
Table \ref{tab:regression}. 
For comparison, we applied the split conformal method \citep{Vovk2005AlgorithmicLI} to the same trained DNNs, \textcolor{black}{with the residual used as the non-conformity score.}
The comparison shows  
the superiority of the post-StoNet approach in terms of prediction interval lengths.  
The sparsity of the post-StoNet 
mitigates potential overfitting issues suffered by the DNNs, thus enhancing prediction performance.
Applying the conformal method on top of over-fitted models led to 
wider prediction intervals.

\begin{table}[!ht]
 \caption{Average coverage rates and lengths of the prediction intervals produced by different methods on the test sets in 10 random splits of the data, with corresponding standard deviations shown in parentheses.}
\begin{adjustbox}{width=1.0\textwidth}
\begin{tabular}{cccccc} \toprule
        Dataset & N & P &Model & Coverage Rate & Interval length  \\ \midrule
     \multirow{3}{*}{Liver} & \multirow{3}{*}{345} & \multirow{3}{*}{5} & Post-StoNet & 90.14\% (3.54\%) &  9.4931 (0.4371)  \\
     &  & &  \textcolor{black}{Post-Linear} & 88.70\% (4.43\%) & {\bf 9.0380 (1.4178)}  \\
         &  & & Split Conformal & 90.72\% (4.82\%) & 12.7296 (0.9708)  \\
        \hline
             \multirow{3}{*}{QSAR} & \multirow{3}{*}{908} & \multirow{3}{*}{6} & Post-StoNet & 89.01\% (2.23\%) & {\bf   3.1315(0.0973) }  \\
              &  & & \textcolor{black}{Post-Linear} & 87.97\% (2.98\%) & 4.9555 (0.3639)  \\
         &  & & Split Conformal & 88.74\% (3.87\%) & 3.5428 (0.2504)  \\
        \hline
        \multirow{3}{*}{Community} & \multirow{3}{*}{1,994} & \multirow{3}{*}{100} & Post-StoNet & 88.25\% (1.84\%) & {\bf   0.4657 (0.0191)}  \\
        &  & & \textcolor{black}{Post-Linear} & 90.03\% (1.97\%) & 0.7065 (0.0655) \\
         &  & & Split Conformal & 89.05\% (2.13\%) & 0.5065 (0.0283) \\
        \hline
        \multirow{3}{*}{STAR} & \multirow{3}{*}{2,166} & \multirow{3}{*}{39} & Post-StoNet & 90.85\% (1.44\%) & {\bf    840.7099 (33.8295) }  \\
        &  & & \textcolor{black}{Post-Linear} & 89.93\% (1.87\%) & 867.2143 (23.2350)  \\
         &  & & Split Conformal & 90.46\% (1.04\%) & 914.9324 (16.5794) \\
         \hline
        \multirow{3}{*}{Abalone} & \multirow{3}{*}{4,177} & \multirow{3}{*}{8} & Post-StoNet & 90.62\% (0.85\%) & {\bf   7.4187 (0.1416)}  \\
        &  & & \textcolor{black}{Post-Linear} & 90.01\% (1.30\%) & 10.0865 (0.9541)  \\
         &  & & Split Conformal & 90.33\% (1.08\%) & 9.6276 (0.2951)  \\
         \hline
        \multirow{3}{*}{Parkinson} & \multirow{3}{*}{5,875} & \multirow{3}{*}{22} & Post-StoNet & 89.40\% (1.23\%) & {\bf 32.1171 (0.6234)}  \\
        &  & & \textcolor{black}{Post-Linear} & 89.71\% (1.14\%) & 36.1565 (0.8758)  \\
         &  & & Split Conformal & 89.49\% (0.70\%) & 35.9594 (0.9017) \\
        \hline
       \multirow{3}{*}{Power Plant}  &  \multirow{3}{*}{9,568}
       &  \multirow{3}{*}{4}  & Post-StoNet & 90.97\% (0.71\%) &  {\bf  13.2852 (0.1723)}  \\
       &  & & \textcolor{black}{Post-Linear} & 89.57\% (1.06\%) & 56.1365 (2.5901)  \\
         &  &  & Split Conformal & 89.99\% (0.82\%) & 14.5719 (0.2676) \\
        \hline
        \multirow{3}{*}{Bike} & \multirow{3}{*}{10,886} & \multirow{3}{*}{18} & Post-StoNet & 89.84\% (0.86\%) & {\bf    173.5158 (2.1959) }   \\
        &  & & \textcolor{black}{Post-Linear} & 89.77\% (0.95\%) & 548.8517 (30.1937)  \\
         &  & & Split Conformal & 89.75\% (0.77\%) & 182.4721 (5.1792)  \\
        \hline
     \multirow{3}{*}{Protein}    &   \multirow{3}{*}{45,730}
      &  \multirow{3}{*}{9} & Post-StoNet & 89.41\% (0.28\%) & {\bf    13.1319 (0.0494)}\\
      &  & & \textcolor{black}{Post-Linear} & 89.96\% (0.29\%) & 18.8432 (0.9714)  \\
         &  &  & Split Conformal & 90.04\% (0.22\%) & 14.4296 (0.0886)  \\
        \hline
    \multirow{3}{*}{Year} &  \multirow{3}{*}{515,345}
     &  \multirow{3}{*}{90} & Post-StoNet & 90.64\% (0.13\%) & {\bf    29.4272 (0.0923)}\\
     &  & & \textcolor{black}{Post-Linear} & 90.27\% (0.47\%) & 31.2775 (1.5763) \\
         & &  & Split Conformal & 90.01\% (0.10\%) & 32.1068 (0.3726)  \\
    \bottomrule
    \end{tabular}
\end{adjustbox}
\label{tab:regression}
\end{table}

\textcolor{black}{To further demonstrate the advantage of using the StoNet for post-transformation data modeling, we again consider the post-Linear method as a baseline. We use the output of the last hidden layer of the neural network as the explanatory variables, and use half of the calibration data to fit a sparse linear model with LASSO penalty, the regularization parameter of LASSO penalty is selected via a 5-fold cross-validation. Then we apply the split conformal method on the other half of the calibration data to compute threshold of the non-conformity score to construct confidence intervals for the test points. 
The comparison further highlights the superiority of the post-StoNet method in uncertainty quantification for deep learning models. Notably, the post-Linear method often yields wider confidence intervals than the original conformal method, though not consistently. This may be attributed to underfitting of the sparse linear model on the validation set.}

\section{Conclusion} \label{sect:discussion}
We have employed the StoNet as a post-processing tool to \textcolor{black}{quantify the prediction  uncertainty for large-scale deep learning models}, and provided a theoretical guarantee for its validity. 
Our numerical results suggest that the post-StoNet approach   
significantly improves prediction uncertainty quantification for deep learning models compared to the conformal prediction method and other post hoc calibration methods.  

\textcolor{black}{We have also shown that the StoNet effectively bridges the gap between linear models and DNNs, allowing us to adapt theories and methods developed for linear models to deep learning models. Specifically, we have adapted sparse learning theory from linear models to DNNs with the Lasso penalty. Extending the results to other amenable penalties \citep{loh2017support},
such as SCAD \citep{FanL2001} and MCP \citep{Zhang2010}, 
is straightforward.}

 \section*{Acknowledgements}

Liang's research is support in part by the NSF grants DMS-2015498 and DMS-2210819, and the NIH grant R01-GM152717. 
The authors thank the editor, associate editor, and referees for their constructive comments which have led to significant improvement of this paper.

\newpage

\appendix

 \section{Theoretical Proofs}

\subsection{Proof of Lemma \ref{lem:eigen_hidden_layer}}
 \begin{proof} 
For simplicity of notations, we suppress the iteration index $t$.
Let $\tilde{\bY}_{l} = \bb_l + \bw_l\Psi(\bY_{l-1})$ for $l = 2,\dots, h$, and let $\tilde{\bY}_{1} = \bb_1 + \bw_1\bX$. By the definition of the StoNet model (\ref{eq:stonet}), 
$\bY_l$ can be  written as $\bY_l = \tilde{\bY}_{l} + \be_l$ for $l\in \{1,2,\ldots,h\}$.

Since $\sigma_l^2$ has been set to a very small value, we have $\Psi(\bY_l) \approx \Psi(\tilde{\bY}_{l}) + \nabla_{\tilde{\bY}_{l}}\Psi(\tilde{\bY}_{l}) \circ \be_l$, where $\circ$ denotes elementwise product. Then
\begin{equation}
\label{recursive_covariance}
\begin{split}
\bSigma_l &\approx \Var(\mathbb{E} (\Psi(\tilde{\bY}_{l}) + \nabla_{\tilde{\bY}_{l}}\Psi(\tilde{\bY}_{l}) \circ \be_l | \tilde{\bY}_{l} ) ) + \mathbb{E}(\Var(\Psi(\tilde{\bY}_{l}) + \nabla_{\tilde{\bY}_{l}}\Psi(\tilde{\bY}_{l}) \circ \be_l | \tilde{\bY}_{l})) \\
& = \Var(\Psi(\tilde{\bY}_{l})) + \diag\left\{\sigma_{l}^2 \mathbb{E}[\nabla_{\tilde{\bY}_{l}}\Psi(\tilde{\bY}_{l}) \circ \nabla_{\tilde{\bY}_{l}}\Psi(\tilde{\bY}_{l})] \right\}, \\
\end{split}    
\end{equation}
where $\diag\{\bv\}$ with $\bv \in \mathbb{R}^{d}$ denotes a $d \times d$ diagonal matrix with diagonal elements being $\bv$.

By Assumption \ref{ass:3}-(iii), the activation function is bounded. For example, {\it tanh} or {\it sigmoid} is used in the model. By Assumption \ref{ass:1}, there exists some constant $C_1$ such that $\|\bb_l\|_{\infty} < C_1, \|\bw_l\|_{\infty} < C_1$. By Assumption \ref{ass:3}, $\|\bX\|_{\infty}$ is bounded. Therefore, there exists some constant $C_2$ such that for any $L \in \{1,2,\ldots,h\}$,  $\|\tilde{\bY}_{l}\|_{\infty} \leq C_1 + C_1 C_2 $ holds by rescaling $\bX$ by a factor of $\prod_{l=1}^h d_l$. Since both $\Psi(\tilde{\bY}_{l})$ and  $\nabla_{\tilde{\bY}_{l}}\Psi(\tilde{\bY}_{l}) $ are bounded, there exists a constant $\kappa_{\max,l}$ such that 
\[
\phi_{\max}(d_{l,n}|\bSigma_l) \leq \kappa_{\max,l}.
\]
To establish the lower bound, we note that $\|\tilde{\bY}_{l}\|_{\infty} \leq C_1 + C_1 C_2 $. Therefore, for an activation function which has nonzero gradients on any closed interval, e.g., {\it tanh} and {\it sigmoid}, there exists a constant $C_3 > 0$ such that  $\min_{i=1,\dots, d_l}\nabla_{\tilde{\bY}_{l}}\Psi(\tilde{\bY}_{l})_i > C_3$, where $\nabla_{\tilde{\bY}_{l}}\Psi(\tilde{\bY}_{l})_i$ denotes the $i$-th element of  $\nabla_{\tilde{\bY}_{l}}\Psi(\tilde{\bY}_{l})$. Then we can take $\kappa_{\min,l} = \sigma_{l,n}^2 C_3^2$ such that 
\[
\phi_{\min}(d_{l,n}|\bSigma_l) \geq \kappa_{\min,l},
\]
which completes the proof.
\end{proof}

\subsection{Proof of Part (i) of Theorem \ref{thm:stonet-IRO}} 
\begin{proof}
By Lemma \ref{lem:eigen_hidden_layer}, $\Sigma_l^{(t)}$ satisfies the requirements of Theorem 1 of \cite{Meinshausen2009LASSOTYPERO} and Theorem 1 of \cite{huang2008iterated}. Then, by Theorem 1 of \cite{Meinshausen2009LASSOTYPERO} (for linear regression) and Theorem 1 of \cite{huang2008iterated} (for logistic regression),  we have $r_n$ as given in the lemma by summarizing the $l_2$-errors of coefficient estimation for all $\sum_{l=1}^{h+1} d_l$ regression/logistic regressions. Further, by the setting of $(\sigma_{1,n}^2, \ldots, \sigma_{h+1,n}^2)$ as specified in Assumption \ref{ass:3}, we have $r_n\to 0$ as $n\to \infty$. This completes the proof of part (i) of Theorem \ref{thm:stonet-IRO}.
\end{proof}

\subsection{Proof of Part (ii) of Theorem \ref{thm:stonet-IRO}} 

\begin{proof}
Then part (ii) of Theorem \ref{thm:stonet-IRO} directly follows from Theorem 4 of \cite{Liang2018missing} that the estimator $\hat{\btheta}_n^{(t)}$ is consistent when both $n$ and $t$ are sufficiently large.
\end{proof}

\subsection{Proof of Corollary \ref{cor:stonet-IRO}} 
 \begin{proof}
 Let $\hat{\btheta}_{n}^{(t)}$ denote the estimate of $\btheta_n$ at iteration $t$, and let $\btheta_{*}^{(t)}$ denote its ``true'' value at iteration $t$, and 
let $\btheta^*$ denote its true value in the  StoNet. 
By the proof of Theorem 4 of \cite{Liang2018missing} and Theorem 1 of \cite{Meinshausen2009LASSOTYPERO}, for the StoNet with the linear regression output layer, we have
\begin{equation}
\mathbb{E}\|\hat{\btheta}_{n}^{(t)} - \btheta^{*}\| \leq \frac{1}{1-\rho^*} \mathbb{E} \|\hat{\btheta}_{n}^{(t)} - \btheta_{*}^{(t)} \| \prec \frac{\sqrt{r_n}}{1-\rho^*}, \quad \mbox{as $t\to \infty$},
\end{equation}
by summarizing all $d_1+d_2+\cdots+d_{h+1}$ linear regressions, where $\rho^*$ is a constant as defined in Assumption \ref{ass:5}.
For the StoNet with the logistic regression output layer, we have the same result by Theorem 1 of \cite{huang2008iterated}. 
Further, by Markov inequality, there exists a constant $c$ such that  
\[
P\left(\|\hat{\btheta}_{n}^{(t)} - \btheta^{*}\| >  c \sqrt{r_n}\right ) \to 0, \quad \mbox{as $n\to \infty$ and $t\to \infty$}.
\]

Then, by Assumption \ref{ass:6}, 
\begin{itemize}
\item For any $i \in \bgamma^*$, $ \|\hat{\btheta}_{n}^{(t)} - \btheta^{*}\| \leq c \sqrt{r_n}$ implies $|\hat{\btheta}_{i,n}^{(t)}| > 
c \sqrt{r_n}$.
\item For any $i \notin \bgamma^*$, $ \|\hat{\btheta}_{n}^{(t)} - \btheta^{*}\|\leq c  \sqrt{r_n}$ implies  $|\hat{\btheta}_{i,n}^{(t)}| < 
c \sqrt{r_n}$.
\end{itemize}
Therefore, 
\begin{equation}
P(\hat{\bgamma} = \bgamma^*) \geq P((\|\hat{\btheta}_n^{(t)} - \btheta^*\| \leq  c \sqrt{r_n}) \rightarrow 1, \quad \mbox{as $n\to \infty$ and $t\to \infty$}, 
\end{equation}
which concludes the proof.
 \end{proof}

\subsection{Proof of Corollary \ref{cor2:CI}} 
\begin{proof}
This proof involves several notations, including $\widehat{\bSigma}$, 
$\hat{\btheta}$, and $\varsigma_{h+1,j}$. As noted in the main text,
their dependence on the sample size $n$ is implicit and has been 
depressed for notational simplicity. 
As implied by (\ref{varest})-(\ref{fixedestS0}), we have 
$\widehat{\bSigma}_i \to 0$, $i \in \{1,2,\ldots,h\}$,
as $n \to \infty$. Additionally,  as 
$n \to \infty$, 
\[
\|\mu(\bz,\hat{\btheta})-\mu(\bz,\btheta^*)\| \stackrel{p}{\to} 0, 
\]
for any test point $\bz$, and 
\[
\varsigma_{h+1,j}^{2} - \sigma_{h+1}^2 \stackrel{p}{\to} 0, \quad j \in \{1,2,\ldots,d_{h+1}\}. 
\]
Therefore, the nominal coverage rate $1-\alpha$ is asymptotically 
guaranteed as $n \to \infty$.  
\end{proof}

\section{The Imputation Regularized-Optimization Algorithm for 
StoNet Training}

This algorithm is given in Algorithm \ref{IROforstonet}. 

 \begin{algorithm}
\caption{IRO Algorithm for StoNet} 
\label{IROforstonet}
\KwInput{Dataset $D_n=(\mathbb{Y},\mathbb{X})$, total iteration number $T$,  and Monte Carlo step number $t_{MC}$.}
 
\KwInitialization{Randomly initialize the network parameters $\hat{\btheta}^{(0)}=(\hat{\btheta}_1^{(0)},\ldots, \hat{\btheta}_{h+1}^{(0)})$.}

\For{$t=1,2,\ldots,T$}{ 
 {\bf $\bullet$ Imputation}: For each sample $(\bX^{(i)},\bY^{(i)})$, draw $\bYmis^{(i, t)}$ from $\pi(\bYmis^{(i)}|\bY^{(i)}, \bX^{(i)}$, $\hat{\btheta}_n^{(t-1)}, \bsigma_n^2)$ with a Metropolis or Langevin dynamics kernel by iterating for $t_{MC}$ steps. 

{\bf $\bullet$ Regularized optimization}: Based on the pseudo-complete data $\{(\bY^{(i)},\bYmis^{(i,t)},\bX^{(i)}): i=1,2,\ldots,n\}$, update $\hat{\btheta}_n^{(t-1)}$ by 
   minimizing a penalized loss function, i.e., setting
   \begin{equation} \label{IROeq1}
    \begin{split}
    \hat{\btheta}_n^{(t)} & =\arg\min_{\btheta}\Big\{  -\frac{1}{n} \sum_{i=1}^n \log\pi(\bY^{(i)},\bYmis^{(i, t)}  \big|\bX^{(i)}, \btheta,\bsigma_n^2) 
     + P_{{\lambda}_n}(\btheta)\Big\},
\end{split}
\end{equation}
   where the penalty $P_{{\lambda}_n}(\btheta)$ is chosen such that 
   $\hat{\btheta}_n^{(t)}$ forms a consistent estimator  of 
   \begin{equation} \label{IROeq2}
   \small
   \begin{split}
    & \btheta_*^{(t)}=\arg\max_{\btheta}  \mbE_{\btheta_n^{(t-1)}} \log\pi(\bY,\bYmis |\bX,\btheta,\bsigma_n^2) \\
   & = \arg\max_{\btheta} \int \log \pi(\bYmis, \bY|\bX,\btheta,\bsigma_n^2) 
    \pi(\bYmis|\bY,\bX,\btheta_n^{(t-1)},\bsigma_n^2) \\ 
   & \quad \times \pi(\bY|\bX,\btheta^*,\bsigma_n^2)d\bYmis d\bY,
   \end{split}
   \end{equation}
   where $\btheta_*^{(t)}$ is called the working true parameter at iteration $t$.} 
\KwOutput{$\hat{\theta}_n^{(T)}$}
 \end{algorithm}

\section{Adaptive Stochastic Gradient MCMC for Efficient StoNet Training} \label{AdaptiveSGMCMCsection:App}

The IRO algorithm requires computation on the full dataset at each iteration and, therefore, it is less scalable with respect to big data. In practice, 
we can train the sparse StoNet using the adaptive stochastic gradient MCMC algorithm as proposed in \citep{LiangSLiang2022}. 
To make the paper self-contained, we give a review of the adaptive 
stochastic gradient MCMC algorithm below.

Let $\pi(\bY|\bX,\btheta,\bsigma^2)=\int \pi(\bY,\bYmis|\bX,\btheta,\bsigma^2) d\bYmis$ denote the likelihood function of the observed data for the StoNet model.
By Fisher's identity, we have 
\[
\nabla_{\btheta} \log \pi(\bY|\bX,\btheta,\bsigma^2)= \int \nabla_{\btheta} \log \pi(\bY,\bYmis|\bX,\btheta,\bsigma^2) \pi(\bYmis|\bY,\bX,\btheta,\bsigma^2) d\bYmis,
\]
which implies the sparse StoNet can also be trained by solving the equation 
\begin{equation} \label{rooteq2}
\int \nabla_{\btheta} [\log \pi(\bY,\bYmis|\bX,\btheta,\bsigma^2)+\log P_{\lambda}(\btheta)] \pi(\bYmis|\bY,\bX,\btheta,\bsigma^2) d\bYmis=0,
\end{equation}
where $P_{\lambda}(\btheta)$ denotes a penalty function satisfying Assumption \ref{ass:4}.
By Theorem 1 of \cite{Liang2018missing}, solving  (\ref{rooteq2}) will lead
to the same solution as solving the optimization problem specified below: 
\[
\widehat{\btheta}_n^*= \arg\max_{\btheta} \Big \{ \frac{1}{n} \sum_{i=1}^n \log\pi(\bY^{(i)}, \bYmis^{(i)}|\bX^{(i)}, \btheta,\bsigma^2) +\frac{1}{n} P_{\lambda}(\btheta) \Big\}.
\]

By \cite{Deng2019adaptive} and \cite{LiangSLiang2022}, the equation (\ref{rooteq2}) can be solved using an adaptive stochastic gradient MCMC algorithm, which works by iterating between the following two steps: 
\begin{itemize} 
\item[(a)] ({\it Sampling}) Generate $\bYmis^{(k+1)}$ from a transition kernel induced by a stochastic gradient MCMC algorithm, e.g.,  
stochastic gradient Hamilton Monte Carlo (SGHMC) \citep{SGHMC2014}.
 
 \item[(b)] ({\it Parameter updating}) Set 
 $\btheta^{(k+1)}=\btheta^{(k)}+ \gamma_{k+1} g(\bYmis^{(k+1)},U_{k+1})$,
 where $\gamma_{k+1}$ denotes the step size used in the stochastic approximation procedure. 
 \end{itemize}
 
 The pseudo-code of the adaptive SGHMC algorithm is given by Algorithm \ref{ASGHMC}, where we let $\btheta_i = (\bw_i, \bb_i)$ denote the parameters associated with layer $i$ of the StoNet,   let $(\bY_0^{(s,k)},\bY_{h+1}^{(s,k)})=(\bX^{(s)},\bY^{(s)})$ denote a training sample $s$, and let $\bY_{mis}^{(s,k)}=(\bY_1^{(s,k)},\ldots,\bY_h^{(s,k)})$ denote the latent variables imputed for the training sample $s$ at iteration $k$. Occasionally, we use the notation $\bY_0^{(s,k)}=\bY_0^{(s)}=\bX^{(s)}$ and $\bY_{h+1}^{(s,k)}=\bY_{h+1}^{(s)}=\bY^{(s)}$. 
 This algorithm is called ``adaptive'' as the transition kernel used in step (i) changes with iterations through the working estimate $\btheta^{(k)}$.
  
\begin{algorithm}
\caption{An adaptive SGHMC algorithm for training StoNet}
\label{ASGHMC}
\KwInput{Dataset $(\bX,\bY)$, total iteration number $K$,  Monte Carlo step number $t_{HMC}$, the learning rate sequence $\{\epsilon_{k,i}:t=1,2,\ldots,T;i=1,2,\ldots,h+1\}$,  and the step size sequence $\{\gamma_{k,i}: t=1,2,\ldots,T; i=1,2,\ldots,h+1\}$. } 
 
\KwInitialization{Randomly initialize the network parameters $\hat{\btheta}^{(0)}=(\hat{\btheta}_1^{(0)},\ldots, \hat{\btheta}_{h+1}^{(0)})$.}
 
 \For{$k=1,2,\ldots,K$}{
    {\bf STEP 0: Subsampling}: Draw a mini-batch of data and denote it by $S_k$.
   
   {\bf STEP 1: Backward Sampling}:
    For each observation $s\in S_k$, sample $\bY_i$'s in the order from layer $h$ to layer $1$. More explicitly, we sample $\bY_i^{(s,k)}$ from the distribution 
    \[
    \pi(\bY_i^{(s,k)}|\hat{\theta}_i^{(k-1)}, \hat{\theta}_{i+1}^{(k-1)},\bY_{i+1}^{(s,k)},\bY_{i-1}^{(s,k)})\propto \pi(\bY_{i+1}^{(s,k)}|\hat{\theta}_{i+1}^{(k-1)},\bY_{i}^{(s,k)})\pi(\bY_{i}^{(s,k)}|\hat{\theta}_{i}^{(k-1)},\bY_{i-1}^{(s,k)})
    \]
    by running SGHMC for $t_{HMC}$ steps:\\
    
    Initialize $\bv_i^{(s,0)}={\bf 0}$, and initialize $\bY_i^{(s,k,0)}$ by forward pass of DNN.\\
    
    \For{$l=1,2,\ldots,t_{HMC}$}{
    \For{$i=h, h-1,\ldots,1$}{ 
    \begin{equation}
    \small
    \begin{aligned} 
    \bv_{i}^{(s, k, l)}=&(1-\epsilon_{k, i} \eta_i) \bv_{i}^{(s, k, l-1)}+\epsilon_{k, i} \nabla_{\bY_{i}^{(s, k, l-1)}} \log \pi\left(\bY_{i}^{(s, k, l-1)} \mid \hat{\btheta}_{i}^{(k-1)}, \bY_{i-1}^{(s, k, l-1)}\right) \\ 
    &+\epsilon_{k, i} \nabla_{\bY_{i}^{(s, k, l-1)}} \log \pi\left(\bY_{i+1}^{(s, k, l-1)} \mid \hat{\btheta}_{i+1}^{(k-1)}, \bY_{i}^{(s, k, l-1)}\right)+\sqrt{2 \epsilon_{k, i} \eta} \be^{(s, k, l)}, \\ 
    \bY_{i}^{(s, k, l)}=& \bY_{i}^{(s, k, l-1)}+ \epsilon_{k, i} \bv_{i}^{(s, k, l-1)} ,
    \end{aligned}
    \end{equation}
    where $\be^{s,k,l}\sim N(0,\bI_{d_i})$, $\epsilon_{k,i}$ is the learning rate, and $\eta$ is the friction coefficient. The algorithm is reduced to SGLD when $\epsilon_{k,i}\eta_i \equiv 1$.
  } }
    
    Set $\bY_i^{(s,k)}=\bY_i^{(s,k,t_{HMC})}$ for $i=1,2,\dots,h$.

    {\bf STEP 2: Parameter Update}:
    Update the estimates of $\hat{\btheta}^{(k-1)}= (\hat{\btheta}_1^{(k-1)},\hat{\btheta}_2^{(k-1)},\ldots,\hat{\btheta}_{h+1}^{(k-1)})$ by stochastic gradient descent
  \begin{equation} \label{SGHMCeq001}
  \small
  \hat{\btheta}_i^{(k)} = \hat{\btheta}_i^{(k-1)}+ \gamma_{k,i}\left(\frac{n}{|S_k|} \sum_{s\in S_k}
   \nabla_{\btheta_i} \log \pi(Y_i^{(s,k)}| \hat{\btheta}_i^{(k-1)}, Y_{i-1}^{(s,k)})  - n\nabla_{\btheta_i} P_{\lambda}(\hat{\btheta}_i)\right), 
  \end{equation}
  for $i=1,2,\ldots, h+1$, where $\gamma_{k,i}$ is the step size used for updating $\theta_i$. 
  }
 \KwOutput{$\hat{\theta}_n^{(K)}$}
\end{algorithm}

\section{Covariance of Latent Variables in the StoNet} \label{sect:CI-APP}

Consider the case that we have a regression StoNet trained by the IRO algorithm. In this case, the prediction uncertainty can be quantified by a recursive application of Eve's law. 

Let $\bz$ denote a test point at which the prediction uncertainty needs to be quantified. For simplicity of notation, we suppress the bias term by including it as a special column of the corresponding weight matrix. To indicate the iterative nature of the IRO algorithm, we include the superscript `$t$' in the derivation.
Let $\bZ_i^{(t)}$ denote the imputed latent variable, corresponding to the input $\bz$,
for layer $i$ at iteration $t$.
For convenience, we let $\bZ_0^{(t)}=\bz$ for all $t$.
 Let $\bmu_i^{(t)}$ and $\bSigma_i^{(t)}$ denote, respectively, the mean and covariance matrix of $\bZ_i^{(t)}$.  Let $\bw_{i_j}^{(t)}$ denote the $j$-th row of the weight matrix $\bw_i^{(t)}$, which represents the weights from the neurons of layer $i-1$ to neuron $j$ of layer $i$. By Eve's law, for any layer $i \in \{2, 3, \dots, h+1\}$, we then have   
\begin{equation}
\small
\nonumber
\begin{split}
  \bSigma_{i}^{(t)} & =\mbE (\Var(\bZ_{i}^{(t)}|\bZ_{i-1}^{(t)})) + \Var(\mbE(\bZ_{i}^{(t)}|\bZ_{i-1}^{(t)})) \\
  &= \mbE \diag\Big\{\psi(\bZ_{i-1}^{(t)}))^T \Var(\hat{\bw}_{i_j}^{(t)}) \psi(\bZ_{i-1}^{(t)})): j = 1, \dots d_i\Big\}  + \Var\Big(\mbE(\hat{\bw}_i) \psi(\bZ_{i-1}^{(t)}) \Big)\\
  &=\diag\Big\{tr(\Var(\hat{\bw}_{i_j}^{(t)}))  \Var(\psi(\bZ_{i-1}^{(t)}))) +  (\mbE(\psi(\bZ_{i-1}^{(t)})))^T \\
  & \quad \times \Var(\hat{\bw}_{i_j}^{(t)})
  (\mbE(\psi(\bZ_{i-1}^{(t)}))): j = 1, \dots d_i \Big\}
   + \mbE(\hat{\bw}_i) \Var(\psi(\bZ_{i-1}^{(t)})) (\mbE(\hat{\bw}_i))^T,
\end{split}
\end{equation}
where $\Var(\hat{\bw}_{i_j}^{(t)})$ is calculated by the Lasso+OLS or Lasso+mLS procedure suggested by \cite{Liu2013AsymptoticPO}. 
We refer to Theorem 3 of \cite{Liu2013AsymptoticPO} for asymptotic normality of the non-sparse components of $\hat{\bw}_{i_j}^{(t)}$. For the OLS case, the non-sparse submatrix of $\Var(\hat{\bw}_{i_j}^{(t)})$ is given by 
 \[
 \widetilde{\Var(\hat{\bw}_{i_j}^{(t)})} = \hat{\varsigma}_{i,j}^2 [(\psi(\widetilde{\BY}_{i-1}^{(t))})^T \psi(\widetilde{\BY}_{i-1}^{(t)})]^{-1},
 \]
 where $\psi(\widetilde{\BY}_{i-1}^{(t)})$ is the design matrix of the linear regression  
 \[
 \BY_{i,j}^{(t)}=\psi(\widetilde{\BY}_{i-1}^{(t)}) (\tilde{\bw}_{i_j}^{(t)})^T+\bepsilon_{i,j}
 \]
 selected by Lasso for neuron $j$ of layer $i$ at iteration $t$, $\bepsilon_{i,j}\sim N(0,\varsigma_i^2I_n)$, and $\hat{\varsigma}_{i,j}^2$ denotes the OLS estimator of $\varsigma_i^2$. Here $\BY_{i-1}^{(t)} \in \mathbb{R}^{n\times d_{i-1}}$ denotes imputed latent variables
 for all neurons of layer $i-1$,  
 $\BY_{i,j}^{(t)} \in \mathbb{R}^n$ denotes imputed latent variables for neuron $j$ of layer $i$, $\widetilde{\BY}_{i-1}^{(t)} \in \mathbb{R}^{n \times \tilde{q}_{i,j}}$ denotes the variables selected by Lasso, $\tilde{\bw}_{i_j}^{(t)}$ denotes the corresponding regression coefficients, 
 and $\tilde{q}_{i,j}$ denotes the number of selected variables.

Let $\bmu_{i-1}^{(t)}=(\mu_{i-1,1}^{(t)}$,  
 $\ldots$,  $\mu_{i-1,d_{i-1}}^{(t)})^T$ denote the mean of $\bZ_{i-1}^{(t)}$, and let  
  $D_{\psi'}(\bmu_{i-1}^{(t)})= \diag\{\psi'(\mu_{i-1,1}^{(t)})$, $\ldots$, $\psi'(\mu_{i-1,d_{i-1}}^{(t)})\}$, 
  where $\psi'$ denotes the first derivative of the activation function $\psi$.
  By the first order Taylor expansion,  we have 
  \[
  \begin{split}
  \mbE(\psi(\bZ_{i-1}^{(t)}))  \approx \psi(\bmu_{i-1}^{(t)}), \quad 
  \Var(\psi(\bZ_{i-1}^{(t)}))  \approx
  D_{\psi'}(\bmu_{i-1}^{(t)}) \Sigma_{i-1}^{(t)} 
  D_{\psi'}(\bmu_{i-1}^{(t)}).
  \end{split} 
  \]
 Further, if we estimate $\mbE(\hat{\bw}_i) $ by $\hat{\bw}_i$ and estimate $\bmu_{i-1}^{(t)}$ by $\bZ_{i-1}^{(t)}$, then  we have the approximation: 
 \begin{equation} \label{varest}
 \small
   \begin{split}
   \widehat{\bSigma}_{i}^{(t)} & \approx
   \diag\Big\{   tr\big( \Var(\hat{\bw}_{i_j}^{(t)}) D_{\psi'}(\bZ_{i-1}^{(t)}) \widehat{\Sigma}_{i-1}^{(t)}  D_{\psi'}(\bZ_{i-1}^{(t)})   \big)  
    + (\psi(\bZ_{i-1}^{(t)}))^T  \Var(\hat{\bw}_{i_j}^{(t)})
   \psi(\bZ_{i-1}^{(t)}): j=1,\ldots,d_i \Big\} \\
   &  +  \hat{\bw}_{i}^{(t)}  D_{\psi'}(\bZ_{i-1}^{(t)}) \widehat{\Sigma}_{i-1}^{(t)} 
  D_{\psi'}(\bZ_{i-1}^{(t)}) (\hat{\bw}_{i}^{(t)})^T.  
   \end{split}
   \end{equation}
For the first hidden layer, it is reduced to  
   \begin{equation} \label{first_layer_varest}
   \begin{split}
   \widehat{\bSigma}_{1}^{(t)} & \approx
   \diag\Big\{ tr\big(  \Var(\hat{\bw}_{1_j}^{(t)}) \Var(\bz) \big)
    + \bz^T  \Var(\hat{\bw}_{1_j}^{(t)}) \bz: 
      j=1,\ldots,d_1 \Big\}  \\
   & \quad  +  \hat{\bw}_{1}^{(t)}  \Var(\bz) (\hat{\bw}_{1}^{(t)})^T.  
   \end{split}
   \end{equation}
Since $\Var(\bz) = 0$ holds for any fixed test point $\bz$,  $\widehat{\bSigma}_{1}^{(t)}$ can be further reduced to 
\begin{equation} \label{fixedestS0}
 \widehat{\bSigma}_{1}^{(t)} \approx
   \diag\left\{ \bz^T  \Var(\hat{\bw}_{1_j}^{(t)}) \bz: j=1,2,\ldots,d_1 \right\}. 
\end{equation}

\subsection{More Numerical Results}

 \begin{table}[!ht]
\caption{Calibration results for CIFAR10 data, where the standard deviations of the respective measures are given in parentheses.}
\centering
\begin{adjustbox}{width=1.0\textwidth}
\begin{tabular}{ccccccc}
\toprule
    Network & Size & Method & ACC(\%) & NLL & ECE(\%)  & CECE(\%)\\
    \midrule
  \multirow{6}{*}{DenseNet40}    &  \multirow{7}{0.75cm}{176K} 
        & No Calibration & {\bf 92.80  (0.08)} & 0.3101 (0.0045) & 4.45  (0.15) & 0.95  (0.03)\\
       &  & Temp. Scaling & {\bf 92.80  (0.08)} & 0.2205 (0.0017) & 1.23  (0.09) & 0.43  (0.02)\\
      &  & Matrix Scaling & 92.36  (0.15) & 0.2277 (0.0034) & 1.28  (0.17) & 0.41  (0.02)\\
    &  & Focal & 92.04  (0.15)  & 0.2377 (0.0037) & 1.36  (0.09) & 0.43  (0.03)\\
    &  & MMCE & 92.24  (0.49)  & 0.2362 (0.0258) & 1.37  (0.35) & 0.44  (0.07)\\
    &  & \textcolor{black}{Post-Linear} & 92.34 (0.25) & 0.2320 (0.0106) & 1.04 (0.99) & 0.44 (0.17) \\
      &  & Post-StoNet & 92.63  (0.13) & {\bf 0.2214 (0.0044)} & {\bf 0.54  (0.07)} & {\bf 0.31  (0.05)} \\
   \midrule
  \multirow{7}{*}{ResNet110}     &  \multirow{7}{0.75cm}{1.7M} 
    & No Calibration & {\bf 92.70  (0.90)} & 0.3359 (0.0472) & 4.84  (0.63) & 1.02  (0.13) \\
      &  & Temp. Scaling & {\bf 92.70  (0.90)}  & 0.2238 (0.0267) & 1.29  (0.11) & 0.45  (0.03) \\
     &  & Matrix Scaling & 92.15  (0.42) & 0.2377 (0.0096) & 1.56  (0.14) & 0.46  (0.02) \\
    &  & Focal & 91.97  (0.29)  & 0.2399 (0.0107) & 0.87  (0.11) & 0.44  (0.03) \\
    &  & MMCE &  91.81  (0.38) & 0.2476 (0.0216) & 1.57  (0.25) & 0.49  (0.07) \\
    &  & \textcolor{black}{Post-Linear} &  92.59 (0.44) & 0.2230(0.0108) & 1.12 (0.29) & 0.39 (0.03)\\
 &  & Post-StoNet & 92.66  (0.84) & {\bf 0.2210 (0.0269)} & {\bf 0.47  (0.23)} & {\bf 0.32  (0.06)} \\
    \midrule
  \multirow{6}{*}{WideResNet-28-10}  & \multirow{7}{0.75cm}{36M} 
  & No Calibration & {\bf 95.84  (0.18)} & 0.1704 (0.0037) & 2.55  (0.09) & 0.56  (0.01) \\
       &  & Temp. Scaling & {\bf 95.84  (0.18)} & 0.1468 (0.0029) & 1.16  (0.05) & 0.34  (0.02) \\
      &  & Matrix Scaling & 93.67  (0.81) & 0.1961 (0.0218) & 1.47  (0.18) & 0.41  (0.02) \\
    &  & Focal &  95.43  (0.07) & 0.1943 (0.0149) & 6.29  (1.33) & 1.37  (0.28) \\
    &  & MMCE &  93.68  (0.81) & 0.1993 (0.0236) & 1.43  (0.09) & 0.46  (0.03) \\
    &  & \textcolor{black}{Post-Linear} & 95.77 (0.13) & {\bf 0.1444 (0.0060)} & 0.94 (0.40) & 0.30 (0.10) \\
     &  & Post-StoNet & 95.64  (0.12) & { 0.1449 (0.0018)} & {\bf 0.87  (0.11)} & {\bf 0.25  (0.02)} \\
\bottomrule
\end{tabular}
\end{adjustbox}
\label{post_stonet_cifar10}
\end{table}

\section{Hyper-parameter Settings for the Numerical Experiments} \label{parametersection}

For Algorithm \ref{ASGHMC}, since the learning rates $\epsilon_{k,i}$'s and the latent variable variances $\sigma_i^2$'s can be largely canceled at each step of latent variable imputation, their absolute values do not mean much to the convergence of the simulation.  
For this reason, we often set their values to be very small in our numerical experiments, which merely controls the size of random noise added to the corresponding  latent variables.

\subsection{Settings for the Illustrative Example} \label{sect:settingsimu}

{\it One-hidden-layer StoNet:} 
we tried three parameter settings: 
\begin{itemize}
\item[(i)] $\sigma_2^2 = 5e-5$, $\sigma_1^2 = 5e-6$, 
$\epsilon_{k,1} = 5e-9$, $\eta_i = \frac{1}{\epsilon_{k,i}}$, $t_{HMC} = 1$, $\frac{\gamma_{k,1}}{|S_k|} = 5e-4$, $\frac{\gamma_{k,2}}{|S_k|} = 5e-8$,  $|S_k|=50$;
\item[(ii)] $\sigma_2^2 = 1e-4$, $\sigma_1^2 = 1e-5$, $\epsilon_{k,1} = 1e-8$, $\eta_i = \frac{1}{\epsilon_{k,i}}$,$t_{HMC} = 1$, $\frac{\gamma_{k,1}}{|S_k|} = 5e-4$, $\frac{\gamma_{k,2}}{|S_k|} = 5e-8$, $|S_k|=50$; 
\item[(iii)] $\sigma_2^2 = 2e-4$, $\sigma_1^2 = 2e-5$, 
$\epsilon_{k,1} = 2e-8$, $\eta_i = \frac{1}{\epsilon_{k,i}}$,$t_{HMC} = 1$, $\frac{\gamma_{k,1}}{|S_k|} = 5e-4$, $\frac{\gamma_{k,2}}{|S_k|} = 5e-8$, $|S_k|=50$.
\end{itemize}

{\it Two-hidden-layer StoNet:} we tried three parameter settings: 
\begin{itemize}
\item[(i)] $\sigma_3^2 = 5e-10, \sigma_2^2 = 5e-11, \sigma_1^2 = 5e-12$, $\epsilon_{k,2} = 5e-14, \epsilon_{k,1} = 1e-14$, $\eta_i = \frac{1}{\epsilon_{k,i}}$,$t_{HMC} = 1$, $\frac{\gamma_{k,3}}{|S_k|} = 5e-6, \frac{\gamma_{k,2}}{|S_k|} = 5e-10, \frac{\gamma_{k,1}}{|S_k|} = 5e-14$, $|S_k|=50$;
\item[(ii)] $\sigma_3^2 = 1e-9, \sigma_2^2 = 1e-10, \sigma_1^2 = 1e-11$, $\epsilon_{k,2} = 1e-13, \epsilon_{k,1} = 1e-14$, $\eta_i = \frac{1}{\epsilon_{k,i}}$,$t_{HMC} = 1$, $\frac{\gamma_{k,3}}{|S_k|} = 5e-6, \frac{\gamma_{k,2}}{|S_k|} = 5e-10, \frac{\gamma_{k,1}}{|S_k|} = 5e-14$, $|S_k|=50$;
 \item[(iii)] $\sigma_3^2 = 2e-9, \sigma_2^2 = 2e-10, \sigma_1^2 = 2e-11$, $\epsilon_{k,2} = 1e-13, \epsilon_{k,1} = 1e-14$, $\eta_i = \frac{1}{\epsilon_{k,i}}$,$t_{HMC} = 1$, $\frac{\gamma_{k,3}}{|S_k|} = 5e-6, \frac{\gamma_{k,2}}{|S_k|} = 5e-10, \frac{\gamma_{k,1}}{|S_k|} = 5e-14$,$|S_k|=50$.
 \end{itemize}

 For both StoNets, the major difference among the settings is at $\sigma_i$'s. For convenience, we call the settings (i), (ii) and (iii), respectively. 

 \subsection{Settings for the Experiments in Section \ref{sect:examples}}
 
 {\it CIFAR100 and CIFAR10:} Following the setting of post-calibration methods in \cite{guo2017calibration}, we split the training data into a training set of 45,000 images and a holdout validation set of 5,000 images. The training settings for the three models are:
 \begin{itemize}
     \item {\it ResNet110}: The model was trained on the training set using SGD with momentum for 200 epochs with the batch size 128, momentum 0.9, and weight decay 0.0001. The learning rate was set to 0.1 for the first 80 epochs and divided by 10 at the 80-th and 150-th epochs.
     \item {\it Densenet40}: The model was trained on the training set using SGD with momentum for 300 epochs with the batch size 128, momentum 0.9, and weight decay 0.0001. The learning rate was set to 0.1 for the first 150 epochs and divided by 10 at the 150-th and 225-th epochs.
     \item {\it WideResNet-28-10}: The model was trained on the training set using SGD with momentum for 200 epochs with momentum 0.9, and weight decay 0.0005. The learning rate was set to 0.1 for the first 60 epochs and divided by 10 at 60-th, 120-th, and 160-th epochs.
 \end{itemize}
 After training, we extracted the outputs of the last fully connected layer of each model on the validation set, and used them as input to a StoNet model with one hidden layer, 100 hidden units, and  the activation function {\it tanh}. The StoNet model was trained using Algorithm (\ref{ASGHMC}) with the hyper-parameters as given in Table \ref{cifar10_post_stonet}. The regularization parameter $\lambda$ was set to  $1e-4$ for CIFAR10 and $5e-5$ for CIFAR100. 
 As a baseline, we consider the Post-Linear model. We used the outputs of the last fully connected layer of each model as input features, and trained sparse multi-class logistic regression models with LASSO penalty on the validation set. The regularization parameter is selected via a 5-fold cross-validation using the default setting in the {\it scikit-learn} package.

\begin{table}[htbp]
    \caption{Post-StoNet Hyper-Parameter Setting for CIFAR10 and CIFAR100 data}
\centering
\vspace{2mm}
    \begin{tabular}{cc}
    \toprule
    Hyper-Parameter & Value \\
    \midrule
        $[\sigma_1^2, \sigma_2^2] $ &  [1e-2, 1e-3] \\
        \hline
         $\epsilon_{k,1}$ & 1e-7 \\
         \hline
         $\eta_1$  & $\frac{1}{\epsilon_{k,1}}$ \\
         \hline
         $t_{HMC}$ & 1 \\
         \hline
         $[\gamma_{k,1}, \gamma_{k,2}]$ & $[\frac{5e-4}
         {5000}, \frac{5e-6}{5000}]$\\
         \hline
         $|S_k|$ & 50\\
         \hline
         $P_{\lambda}(\btheta)$ & $\lambda \|\btheta\|_1 $\\
    \bottomrule
    \end{tabular}
    \label{cifar10_post_stonet}
\end{table}

{\it Regression Examples:} The data sets were from UCI machine learning repository. For all experiments, we split the data into 40\% as the training set, 40\% as the calibration/validation  set (used to fit a StoNet model for our approach and to compute the absolute value of the residue as the non-conformity score for Split Conformal), and 20\% as the test set. The random split was repeated 10 times. We reported the mean values and standard deviations of the prediction interval length and coverage rate. 

We modeled each dataset using a DNN with 2 hidden layers, with 1000 and 100 hidden units respectively, and the activation function {\it tanh}. The DNN was trained using Adam \citep{adam2015} with a batch size of 50 and a constant learning rate of 0.001. 
The algorithm was run for 1000 epochs for the Protein data set, 200 epochs for the Year data set, and 5000 epochs for other data sets. After the DNN was trained, we refit a StoNet on the calibration set using the output of the last hidden layer of the DNN as input. The StoNet had one hidden layer, 20 hidden units, and the  activation function {\it tanh}. Algorithm \ref{ASGHMC} was used to train the StoNet with the hyper-parameters as given in Table \ref{uci_post_stonet}. The penalty parameter $\lambda$ was selected from $\{1e-1, 8e-2, 5e-2, 4e-2, 3e-2, 2e-2, 1e-2, 5e-3, 2e-3, 1e-3, 5e-4\}$ by 5-fold cross-validation, where we picked a value of $\lambda$ such that the average coverage rate on the calibration sets were closest to the target level 90\%. Specifically, we picked  $\lambda = 8e-2$ for the Liver dataset\footnote{\url{https://archive.ics.uci.edu/dataset/60/liver+disorders}}, $\lambda = 8e-2$ for QSAR dataset\footnote{\url{https://archive.ics.uci.edu/dataset/504/qsar+fish+toxicity}}, $\lambda = 3e-3$ 
for Community dataset\footnote{\url{https://archive.ics.uci.edu/dataset/183/communities+and+crime}},
$\lambda = 8e-2$ for STAR dataset\footnote{\url{https://github.com/yromano/cqr/tree/master/datasets}},
$\lambda = 5e-2$ for Abalone dataset\footnote{\url{https://archive.ics.uci.edu/dataset/1/abalone}},
$\lambda = 5e-2$ for Parkinson dataset\footnote{\url{https://archive.ics.uci.edu/dataset/189/parkinsons+telemonitoring}},
$\lambda = 5e-3$ for Power
Plant dataset\footnote{\url{https://archive.ics.uci.edu/dataset/294/combined+cycle+power+plant}},
$\lambda = 2e-3$ for Bike dataset\footnote{\url{https://archive.ics.uci.edu/ml/datasets/bike+ sharing+dataset}},
$\lambda = 5e-3$ for Protein dataset\footnote{\url{https://archive.ics.uci.edu/dataset/265/physicochemical+properties+of+protein+tertiary+structure}},
$\lambda = 1e-3$ for Year dataset\footnote{\url{https://archive.ics.uci.edu/dataset/203/yearpredictionmsd}}

  \begin{table}[htbp]
  \centering
\caption{StoNet Hyper-Parameter Setting for UCI data sets, where $N$ is size of the calibration data set.}
\vspace{2mm}
\begin{tabular}{cc} \toprule
    Hyper-Parameter & Value \\
    \midrule
    $[\sigma_1^2, \sigma_2^2] $ &  [1e-4, 1e-5] \\
    \hline
     $\epsilon_{k,1}$ & 1e-7 \\
     \hline
     $\eta_1$  & $\frac{1}{\epsilon_{k,1}}$ \\
     \hline
     $t_{HMC}$ & 1 \\
     \hline
     $[\gamma_{k,1}, \gamma_{k,2}]$ & $[\frac{1e-3}
     {N}, \frac{1e-5}{N}]$\\
     \hline
     $|S_k|$ & 50\\
     \hline
     $P_{\lambda}(\btheta)$ & $\lambda  \|\btheta\|_1 $\\
\hline
\end{tabular}
\label{uci_post_stonet}
\end{table}

\section{Consistency is Essential for the Validity of the Post-StoNet Approach} \label{sect:non-consistency}

The parameter estimation consistency is essential for the validity of the post-StoNet approach. 
To demonstrate this issue,
we applied the post-StoNet modeling approach to 
 the Community dataset without regularization (i.e. setting $\lambda = 0$), which violates 
 the sparsity condition of Theorem \ref{thm:stonet-IRO}. The resulting prediction intervals have 
 only a coverage rate of $29.25\%$ (with a standard deviation of $3.77\%$).

\bibliographystyle{chicago}
\bibliography{reference}

\end{document}